\documentclass[pdflatex,sn-nature,iicol,oneside]{sn-jnl}

\usepackage[utf8]{inputenc}
\usepackage{graphicx}%
\usepackage{listings}
\usepackage{multirow}%
\usepackage{amsmath,amssymb,amsfonts}%
\usepackage{amsthm}%
\usepackage{mathrsfs}%
\usepackage[title]{appendix}%
\usepackage{xcolor}%
\usepackage{textcomp}%
\usepackage{manyfoot}%
\usepackage{booktabs}%
\usepackage{algorithm}%
\usepackage{algorithmicx}%
\usepackage{algpseudocode}%
\usepackage{listings}%
\usepackage{geometry}
\usepackage{bm}%
\usepackage{comment}
\usepackage{longtable, booktabs}

\theoremstyle{thmstyleone}%
\theoremstyle{thmstyletwo}%

\theoremstyle{thmstylethree}%

\begin{document}

\title[Article Title]{Predicting and generating antibiotics against future pathogens with ApexOracle}

\author[1-2]{\fnm{Tianang} \sur{Leng}}\email{ltianang@seas.upenn.edu}
\equalcont{Equal Contribution}

\author[1-4]{\fnm{Fangping} \sur{Wan}}\email{fangping.wan@pennmedicine.upenn.edu}
\equalcont{Equal Contribution}

\author[1-4]{\fnm{Marcelo} \sur{Der Torossian Torres}}

\author*[1-4]{\fnm{Cesar} \sur{de la Fuente-Nunez}}\email{cfuente@upenn.edu}

\affil[1]{\orgdiv{Machine Biology Group, Departments of Psychiatry and Microbiology, Institute for Biomedical Informatics, Institute for Translational Medicine and Therapeutics, Perelman School of Medicine}, \orgname{University of Pennsylvania}
}

\affil[2]{\orgdiv{Departments of Bioengineering and Chemical and Biomolecular Engineering, School of Engineering and Applied Science}, \orgname{University of Pennsylvania} \orgaddress{\city{,Philadelphia}, \state{Pennsylvania}, \country{United States of America}}}

\affil[3]{\orgdiv{Department of Chemistry, School of Arts and Sciences}, \orgname{University of Pennsylvania} \orgaddress{\city{,Philadelphia}, \state{Pennsylvania}, \country{United States of America}}}

\affil[4]{\orgdiv{Penn Institute for Computational Science}, \orgname{University of Pennsylvania} \orgaddress{\city{,Philadelphia}, \state{Pennsylvania}, \country{United States of America}}}

\abstract{Antimicrobial resistance (AMR) is escalating and outpacing current antibiotic development. Thus, discovering antibiotics effective against emerging pathogens is becoming increasingly critical. However, existing approaches cannot rapidly identify effective molecules against novel pathogens or emerging drug-resistant strains. Here, we introduce ApexOracle, an artificial intelligence (AI) model that both predicts the antibacterial potency of existing compounds and designs de novo molecules active against strains it has never encountered. Departing from models that rely solely on molecular features, ApexOracle incorporates pathogen-specific context through the integration of molecular features—captured via a foundational discrete diffusion language model—and a dual-embedding framework that combines genomic- and literature-derived strain representations. Across diverse bacterial species and chemical modalities, ApexOracle consistently outperformed state-of-the-art approaches in activity prediction and demonstrated reliable transferability to novel pathogens with little or no antimicrobial data. Its unified representation–generation architecture further enables the \textit{in silico} creation of “new-to-nature” molecules with high predicted efficacy against priority threats. By pairing rapid activity prediction with targeted molecular generation, ApexOracle offers a scalable strategy for countering AMR and preparing for future infectious-disease outbreaks.
}

\keywords{Deep Learning, Diffusion Language Model, Generative AI, Representation learning, Drug Discovery, Antibiotics, Antimicrobials}

\maketitle
\clearpage

\begin{figure*}[htbp]
  \centering
  \includegraphics[width=\textwidth]{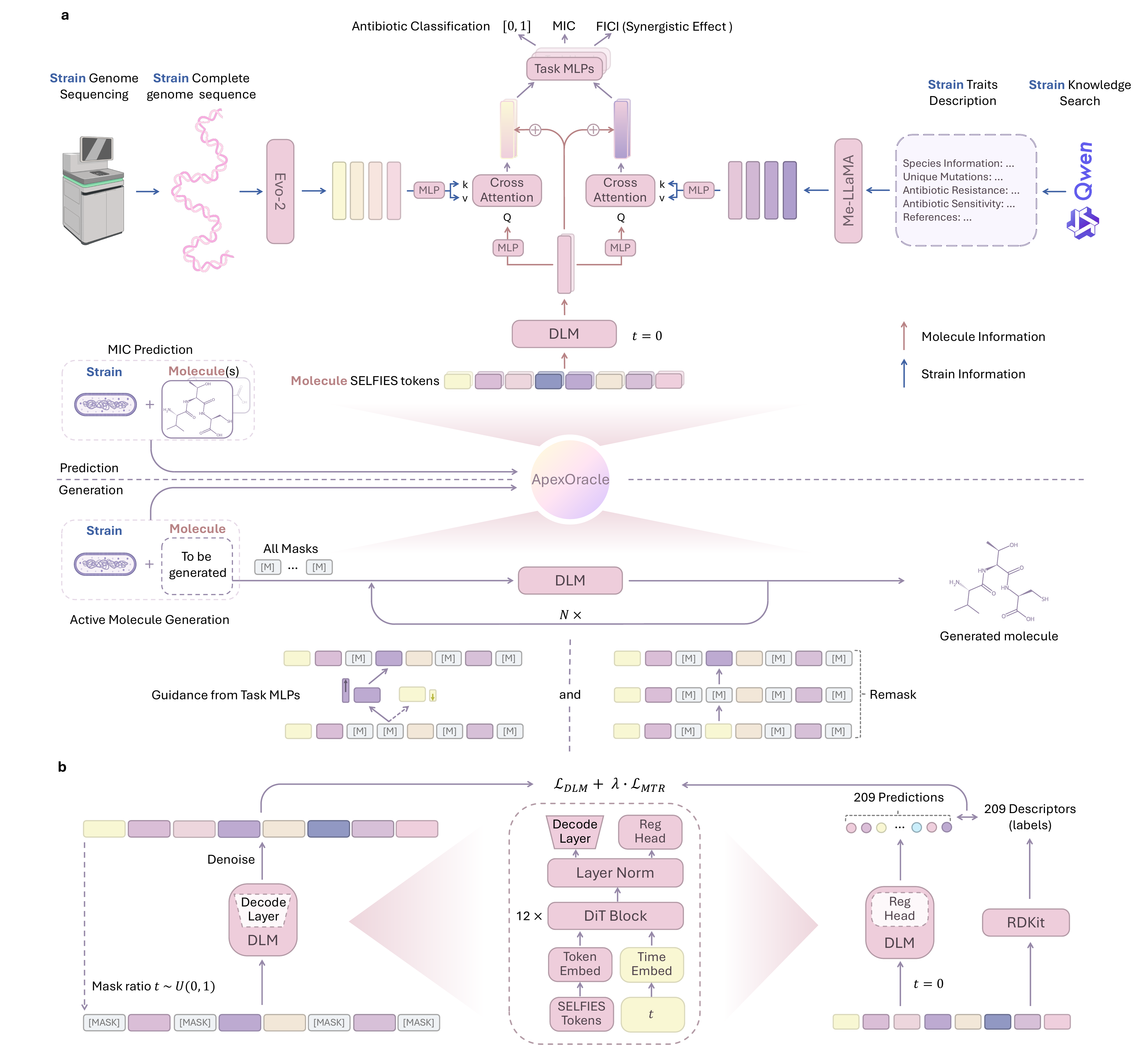}
  \caption{\textbf{The architecture of ApexOracle and DLM training tasks. a.} The architecture of ApexOracle, a unified generative model that can do both pathogen strain knowledge-aware antimicrobial activity prediction and \emph{de novo} generation of active antimicrobial compounds. \textbf{b.} Our DLM training regimen: the model is trained not only to reconstruct molecules from scratch, but also to perform multi-target regression—learning high-quality latent embeddings that link specific token sequences to molecular properties. The 209 target descriptors were computed by RDKit.}
  \label{fig:all}
\end{figure*}
\section{Introduction}\label{sec1}

The rise of antibiotic-resistant bacteria is a global health crisis, making once-curable infections increasingly deadly. Recent estimates attribute approximately 4.95 million deaths per year to illnesses involving drug-resistant pathogens\cite{o2014antimicrobial}. The pipeline of new antibiotics has not kept pace with this threat, in part because traditional drug discovery is painstakingly slow and expensive. Developing a single new antibiotic candidate typically spans many years of lab work and clinical testing, and even promising leads can fail due to unanticipated resistance or toxicity. Compounding the problem, future pathogens, whether emerging natural strains or engineered organisms, may have resistance mechanisms or biological traits for which no known antibiotics are effective. There is a critical need for innovative approaches that can rapidly identify or design effective antibiotics against both known and yet-to-emerge pathogens.

Artificial intelligence has recently shown promise in accelerating antimicrobial discovery\cite{wong2023leveraging, maasch2023molecular, santos2024discovery, APEX, torres2022mining}. Machine learning models can mine large datasets to predict antimicrobial activity and navigate chemical space far more efficiently than conventional methods. Nevertheless, most advances tackle only fragments of the discovery pipeline. For instance, the majority of prior efforts \cite{xu2021comprehensive, sidorczuk2022benchmarks, wang2024diff, xing2024iamp, cao2025tg, chen2024tp} focus exclusively on predicting general antimicrobial peptides (AMPs) without accounting for pathogen-specific activity. More advanced frameworks, such as APEX \cite{APEX}, can predict AMP efficacy against multiple predetermined strains, yet still struggle to generalize to novel pathogens because of their fixed target sets and the scarcity of minimum inhibitory concentration (MIC) data. Leading small-molecule antibiotic prediction models \cite{wong2024discovery, liu2023deep} face a similar limitation: they are strain-specific and require large, bespoke datasets to train separate models for each bacterial target—a process that is labor-intensive and expensive. A recent study \cite{chung2024ensemble} attempted to overcome this by embedding each genome into an 84-dimensional vector of concatenated k-mer composition profiles; however, this strategy effectively erases gene-level signals, produces non-interpretable representations, and was evaluated on only three strains, leaving its broader applicability untested. Overall, these approaches employ separate peptide-only or small-molecule-only models, restricting cross-modal generalization and leaving valuable data under-utilized. Generation models~\cite{jin2025ampgen,wang2024diff,cao2025tg} suffer from the same limitation: their targeting ability is constrained by the predictor that guides them. Therefore, a unified approach that can leverage heterogeneous molecular data (e.g., small molecules and peptides) and generalize across diverse bacterial strains is urgently needed. 

To bridge these gaps, we developed ApexOracle (Fig.\hyperref[fig:all]{1}), a unified machine learning platform for pathogen knowledge-aware antimicrobial discovery. Oracle stands for Omnimodel for pRedictive and generAtive antimicrobiaL discovEry. ApexOracle uniquely integrates multimodal inputs, including pathogen genomic and textual knowledge together with molecular features, via cross-attention mechanisms that capture intricate relationships among a pathogen’s genetic profile, phenotypic traits, and candidate drug structures. Specifically, the model fuses: i) a diffusion language model (DLM) that serves as a joint representation learning and generative engine for compounds; ii) Evo2~\cite{Evo2}, which extracts pathogen genomic embeddings; and iii) Me‑LLaMA~\cite{Me-llama} for semantic embedding extraction from pathogen-related knowledge generated by Qwen2.5-Max~\cite{qwen25}.

By incorporating pathogenic context, ApexOracle is explicitly designed to generalize to future or previously unseen pathogens. Given a new genome sequence or phenotypic description, the model can infer likely drug susceptibilities and even propose entirely new molecules predicted to neutralize the threat. Trained on the SELFIES~\cite{krenn2020self} chemical representation, ApexOracle seamlessly handles both small molecules and peptides, including those bearing non-canonical amino acids, terminal modifications, or intrachain bonds. This capability is critical as 71.5\% of AMPs in DBAASP~\cite{pirtskhalava2021dbaasp} contain non-canonical elements, yet many conventional predictors assume purely canonical, linear peptides and thus oversimplify the AMP landscape.

We evaluate ApexOracle across a broad spectrum of bacterial and fungal strains and demonstrate that its multimodal design yields state-of-the-art accuracy in predicting antimicrobial outcomes for both small molecules and peptides. Furthermore, by conditioning its generative module on pathogen embeddings and guiding it with the internal activity predictor, ApexOracle designs \textit{de novo} candidate antimicrobials with high predicted potency.  

Collectively, these results outline a new paradigm for antimicrobial discovery. Rather than searching chemical space blindly, we can now predict and design compounds in tandem—guided directly by pathogen blueprint-informed antimicrobial predictor—spanning an expansive chemical landscape. This synergy opens new avenues for accelerating the identification of novel therapeutics and paves the way for proactive responses to evolving resistance. We conclude by discussing the broader implications of ApexOracle for staying ahead of infectious-disease threats and highlighting future enhancements that could further align AI-driven drug design with urgent clinical needs. 

\section{Results}\label{sec2}
\begin{figure*}[htbp]
  \centering
  \includegraphics[width=\textwidth]{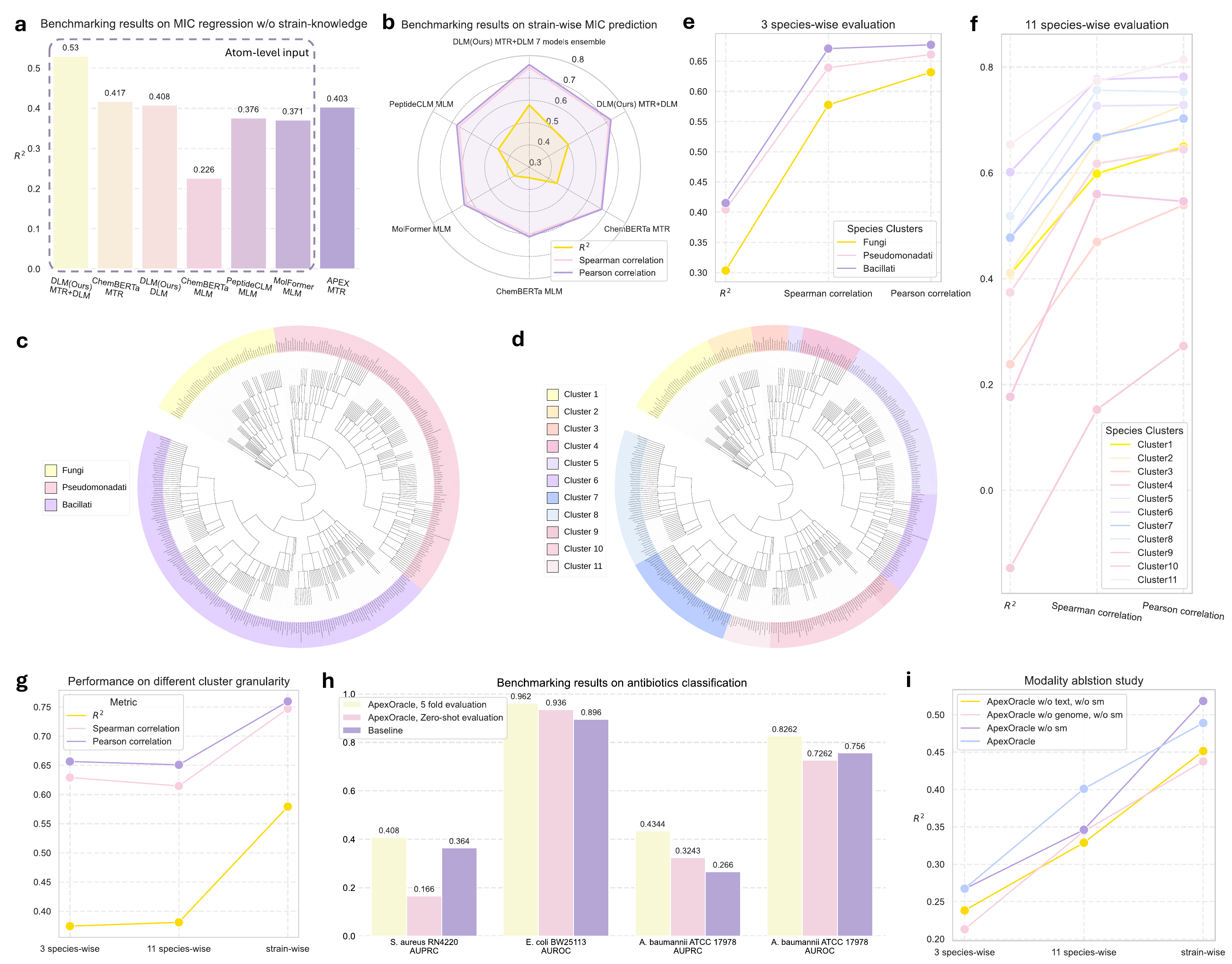}
  \caption{\textbf{Evaluation of ApexOracle prediction performance. a.} Benchmarking molecular representations from our diffusion language model (DLM) against alternatives for pathogen strain-knowledge-unaware MIC regression ($R^2$). All comparators except APEX were language models pretrained solely on atom-level inputs. APEX is a deep neural network that predicts antimicrobial activity from amino acid-level peptide inputs. \textbf{b.} Pathogen strain-knowledge–aware MIC prediction benchmarking under the strain-wise setting: our DLM and its ensemble consistently outperformed rival models. \textbf{c.} Hierarchical clustering of the dataset into three major taxonomic groups: Fungi, Bacillati (Gram-positive), and Pseudomonadati (Gram-negative).  \textbf{d.} Hierarchical clustering into eleven distinct species-level clusters. \textbf{e.} Performance across three major taxonomic divisions (R², Spearman correlation, Pearson correlation). \textbf{f.} Performance across eleven finer species clusters. Robust accuracy was maintained for most clusters; the lone outlier (cluster 9, Mycoplasmatota) likely reflects its atypical biology. \textbf{g.} Impact of cluster granularity on ApexOracle performance. Results indicate that ApexOracle benefits from finer cluster granularity (strain-wise $>$ 11 species-wise $>$ 3 species-wise), while still maintaining moderate antimicrobial prediction performance on phylogenetically distant species. \textbf{h.} Small-molecule antibiotic classification benchmark. Even in zero-shot tests, ApexOracle surpassed baseline methods on two of three strains; with five-fold cross-validation and ensembling, it outperformed all baselines \cite{wong2024discovery, liu2023deep, stokes2020deep}. \textbf{i,} Modality ablation study for ApexOracle. Genome and text (annotation) information are of comparable importance and contribute synergistically to overall prediction performance.}
  \label{fig:2}
\end{figure*}
\subsection{ApexOracle architecture}\label{subsec2-1}

To enable generalization to previously unseen strains, we built ApexOracle as a multimodal deep learning architecture (Fig. 1a). The model accepts three inputs representing a pathogen-drug scenario: (1) the pathogen’s genome sequence, (2) a textual description of the pathogen’s traits, and (3) a molecule (either an existing drug or a candidate structure). From these, ApexOracle produces two kinds of outputs: a predicted efficacy (how well the given antibiotic would work against the pathogen) and/or a generated molecule (a new chemical structure predicted to be effective against a pathogen of interest). The design enables prediction and generation within one unified framework. 

ApexOracle integrates three foundational representation modules. First, the genomic encoder employs Evo2, a DNA language model pretrained on genomes spanning all domains of life, to transform a pathogen’s entire genome into a numerical representation that captures genotypic hallmarks such as resistance determinants, essential gene patterns, and other lineage-specific signals. Second, a textual trait encoder based on a fine-tuned Me-LLaMA model digests brief descriptive passages covering taxonomy, Gram status, morphology, ecological niche, and documented resistance phenotypes; the resulting embedding encodes phenotypic context that may not be evident from sequence alone, for example spore formation or biofilm propensity. Finally, a molecular representation learning and generative module operates on SELFIES (Self-Referencing Embedded Strings), using a Diffusion Language Model (DLM) to map antibiotic structures into latent space or, conversely, to generate novel molecules. Together, these embeddings—genomic, textual, and molecular—provide ApexOracle with a unified, pathogen strain-knowledge-rich foundation for both accurate antimicrobial efficacy prediction and predictor-guided antimicrobial molecule design.

Choosing DLM as the backbone of ApexOracle was driven by the observation that traditional masked language models struggle to generate high-quality samples and that both generative and multi-target regression (MTR) pretraining improves molecular property prediction \cite{liu2023group}\cite{feng2024unigem}\cite{ahmad2022chemberta}. The model was trained not only to reconstruct molecules from scratch but also to perform multi-target regression, learning high-quality latent embeddings that relate specific token sequences to molecular properties. A total of 209 property targets were calculated using RDKit (Fig.\hyperref[fig:all]{1b}). To expose the model to a broad chemical space, we curated a diverse dataset of 121.6 million unique sequences from PubChem~\cite{kim2016pubchem}, SmProt v2.0~\cite{li2021smprot}, UniRef~\cite{suzek2015uniref}, UniProt~\cite{uniprot2019uniprot}, and CycloPS~\cite{duffy2011cyclops, peptideclm} generated modification-rich peptides. 

\subsection{Detailed modification information is crutial for AMP efficacy modeling}

Conventional AMP models often assume that all peptides consist of canonical linear amino acid sequences. However, the DBAASP database indicates that 71.5\% of AMP entries include non-canonical amino acids, terminal modifications, or intramolecule bonds. To determine the importance of these specialized signals, and also evaluate the molecular feature representation quality of ApexOracle’s DLM. We benchmarked the predictive performance of ApexOracle’s DLM against other well-known atom-level input (SELFIES or SMILES strings) chemical language models, as well as against a state-of-the-art residue-level input AMP model, APEX~\cite{APEX}. For all language models evaluated here, we extracted first-token embedding from the last transformer
layer as the molecular features. The extracted features were then coupled with fully connected neural network to make downstream MIC predictions. All comparisons were performed on 19 bacterial strains with more than 700 MIC entries in DBAASP using five-fold cross validation, and under a pathogen strain-knowledge-unaware setting—i.e., models were trained and tested exclusively on a predefined panel of strains, with no ability to generalize to unseen strains. As APEX is limited to canonical linear peptides, noncanonical modifications were ignored, and non-standard amino acids were treated as unknown residues. Due to this oversimplification, APEX was outperformed by three atom-level input models (Fig.\hyperref[fig:2]{2a}). Our DLM, jointly trained under the multi-task regression (MTR) objective, produced the best molecular representations by delivering the best results. Specifically, our DLM surpassed the second-best model by $27.1\%$ in $R^2$, underscoring the pivotal role of atomic-level modification information in accurate AMP activity prediction and demonstrating the substantial benefit in terms of molecular representation learning conferred by the MTR objective.

\subsection{ApexOracle generalizes to unseen strains}\label{subsec2-2}

To simulate a “future pathogen” scenario, we designed three hierarchical evaluation strategies for regression of MICs, each holding out a different level of clusters of strains entirely during training. Unlike the pathogen strain-knowledge-unaware setting, here ApexOracle learns from both molecule and pathogen strain knowledge. Even without exposure to examples from certain strains or species, the model remains capable of accurately predicting antibiotic efficacy against them.

We began with the most evolution-like test, strain-wise evaluation, which probes generalization within a single species. For each species, we conducted three-fold validation: two-thirds of its strains were used for training, and the remaining one-third for testing. ApexOracle achieved an average R2 of 0.5793 with an ensemble of seven models (trained under different random seeds) and 0.5032 without ensembling on held-out unseen strains. We further benchmarked several alternative atom-level models as molecular encoders within ApexOracle and found that our DLM consistently outperformed them in the pathogen strain-knowledge-aware setting; without ensembling, the DLM achieved a 13.5\% higher R2 than the next-best model 
(Fig.\hyperref[fig:2]{2b}).

Beyond strain-wise evaluation, we introduced two increasingly stringent strategies that cluster strains at the species level by taxonomic distance: one with three clusters and another with eleven. In the three-cluster setting, strains were grouped into broad taxonomic categories—Fungi, Pseudomonadota (Gram-negative), and Bacillota (Gram-positive). These were subdivided further in the eleven-cluster setting, with Fungi split into two subgroups, Pseudomonadota into four, and Bacillota into five. We refer to the two strategies as the 3 species-wise (Fig.\hyperref[fig:2]{2c}) and 11 species-wise (Fig.\hyperref[fig:2]{2d}) evaluations, respectively. In each case, one cluster was held out for testing while the others were used for training  (Fig.\hyperref[fig:2]{2e, 2f}). The 3 species-wise setting poses the greatest challenge, since it maximizes the taxonomic distance between training and test data. 

Performance decreased as taxonomic divergence widened, yet ApexOracle retained robust predictive power. Even in the most challenging 3-species-wise setting, the ensemble version of our model achieved an average $R^2$ of 0.3744 (Fig.\hyperref[fig:2]{2g}). In the 11 species-wise evaluation, the average $R^2$ reached 0.4337, excluding the sole exception of cluster 9 (Mycoplasmatota, Fig.\hyperref[fig:2]{2d, 2f}), whose members lack conventional peptidoglycan cell walls and instead possess cholesterol-rich, highly fluid membranes derived from host sterols. 

Altogether, these findings demonstrate ApexOracle’s capacity to extrapolate beyond its training distribution, leveraging genetic and molecular context to generate accurate predictions across previously unseen pathogen clades. This ability to reason mechanistically, rather than merely memorizing strain-specific patterns, is essential for anticipating the behavior of novel or emerging pathogens.

\subsection{ApexOracle outperforms existing methods on small molecule antibiotics discovery without strain-specific labels}\label{subsec2-3}

Beyond its strong capacity in AMP activity prediction, ApexOracle also excels at predicting small molecule antibiotic activity. Previous models typically depend on sizeable, direct training sets for each target strain \cite{wong2024discovery, liu2023deep, stokes2020deep}, but ApexOracle leverages knowledge distilled from tens of thousands of molecules and pathogens—spanning multiple data modalities—to make accurate predictions even when no small molecule antibiotic labels exist for the strain of interest.

To quantify this advantage, we benchmarked ApexOracle against three state-of-the-art small-molecule classifiers \cite{wong2024discovery, liu2023deep, stokes2020deep}, each trained on data measured against a specific pathogen strain \textit{S. aureus} RN4220, \textit{E. coli} BW25113, and \textit{A. baumannii} ATCC 17978, respectively. In a zero-shot setting—without fine-tuning on strain-specific small-molecule data—ApexOracle, trained on AMP data along with the other two test-strain-irrelavant small-molecule antibiotics datasets, matched or outperformed two of the four fine-tuned baseline models (Fig.\hyperref[fig:2]{2h}, pink). 

For a direct comparison with prior work, we then mimicked each study’s evaluation by using five-fold cross-validation with ensemble predictions from ten independently seeded models per fold. Under this setting, fine-tuned ApexOracle decisively outperformed every baseline, delivering average gains of 8.3\% in AUROC and 37.7\% in AUPRC across the three strains (Fig.\hyperref[fig:2]{2h}, yellow). 

These findings demonstrate that the broad molecular-pathogen knowledge captured by ApexOracle transfers effectively to new strains and molecules, providing substantial accuracy improvements without the prohibitive cost of assembling new strain-specific antibiotic dataset.

\begin{figure*}[htbp]
  \centering
  \includegraphics[width=\textwidth]{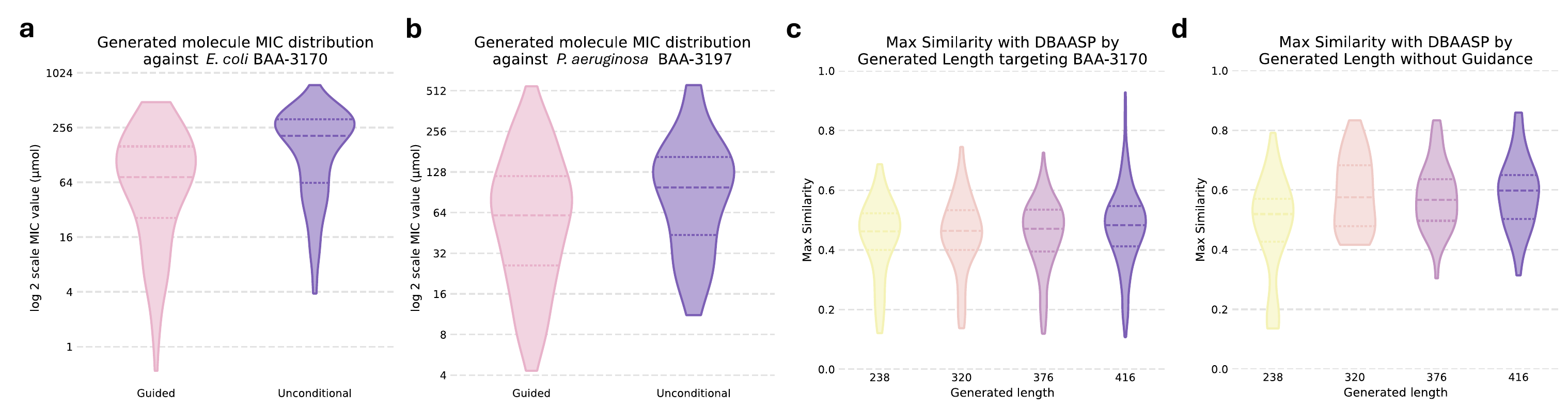}
  \caption{\textbf{Predicted MIC values and novelty of generated molecules. a.} Predicted MIC distributions for molecules generated with versus without pathogen guidance targeting \textit{E. coli} BAA-3170. \textbf{b.} Same comparison for \textit{P. aeruginosa} BAA-3197. \textbf{c.} Maximum Tanimoto similarity of pathogen-guided molecules (binned by token length) to DBAASP compounds, targeting \textit{E. coli} BAA-3170. \textbf{d.} Corresponding similarity for unconditionally generated molecules. Lower similarity in panels c and d indicates greater structural novelty.}
  \label{fig:3}
\end{figure*}

\subsection{Synergistic effect prediction with ApexOracle}

Antibiotic synergy provides a powerful strategy to achieve more potent bacterial killing, reduce the emergence of resistance, and lower toxicity by enabling effective treatment at reduced drug doses. To explore pathogen strain-knowledge-aware antibiotic combination outcome prediction, thereby expanding ApexOracle's applicability, we concatenated the ApexOracle-fused molecular features of two molecules and fed the combined representation into a downstream neural network to predict antimicrobial synergy. This ApexOracle-based synergy model was trained and evaluated using a strain-wise, 3-fold cross-validation methodology, as detailed in Section \ref{subsec2-2}.

The synergy training data, curated from the DBAASP database, encompassed a broader range of compounds than solely AMPs. In total, our curated dataset comprised 2,732 unique molecule synergy-strain pairs, of which 88\% were AMP-small molecule interactions and the remaining 12\% were AMP-AMP interactions. Recognizing the challenges posed by the scarcity and inherent noise in synergy data, we binarized the FICI values using a 0.5 cutoff—labeling values below 0.5 as synergistic (1) and those equal to or above 0.5 as indifferent (0)—and employed an ensemble strategy. By assembling 7 models, we achieved a mean 3-fold strain-wise AUROC of 0.7539 and AUPRC of 0.7454.

\subsection{Modality ablation studies}

To pinpoint which information streams drive ApexOracle’s high accuracy, we performed systematic ablation experiments that removed one modality at a time—genomic features, pathogen‐specific textual descriptions, or a small-molecule antibiotic classification task—and measured the impact on MIC regression performance (Fig.\hyperref[fig:2]{2i}). Ablations were conducted on a curated subset of DBAASP containing 67,304 AMP–strain activity pairs for which both genome and text data were available, ensuring a fair comparison across settings. To accelerate testing while keeping all ablations orthogonal to molecular encoding, we substituted the DLM’s molecule encoder with ChemBERTa, leaving the remainder of the architecture unchanged.

Eliminating any single modality degraded predictive performance, confirming that each contributes meaningfully. Removing genomic embeddings caused the sharpest drop, closely followed by excluding textual annotations, indicating strong synergy between those two pathogen-context channels. Dropping the auxiliary small-molecule classification task produced a smaller but still non-negligible decline, showing that cross-modal auxiliary learning refines molecular representations and improves AMP MIC generalization.

\subsection{Designing novel antibiotics for unseen strains via predictor-guided generation}

Beyond prediction, ApexOracle functions as a \textit{de novo} antimicrobial designer. Its generative module accepts our pathogen knowledge-informed antimicrobial prediction as conditional signals and outputs candidate SELFIES strings predicted to exploit that organism’s specific vulnerabilities.

To test this capacity in a “future-pathogen” scenario, we instructed ApexOracle to propose molecules active against two high-priority strains it had never seen: \textit{E. coli} BAA-3170 (colistin-resistant) and \textit{P. aeruginosa} BAA-3197 (fluoroquinolone-, beta-lactam-, and carbapenem-resistant). Relative to unconditional sampling, our predictor-guided generation produced predicted MIC distributions strongly shifted toward lower values (Fig.\hyperref[fig:3]{3a, 3b}). We then computed each generated molecule’s maximum Tanimoto similarity to compounds in DBAASP—the same dataset that trained the MIC predictor steering generation. Guided molecules showed lower similarity than those from unconditional runs (Fig.\hyperref[fig:3]{3c, 3d}), indicating that ApexOracle is designing genuinely novel chemotypes rather than copying training examples.

\section{Discussion}\label{sec12}
We have introduced ApexOracle, an AI-driven platform that couples pathogen genomics with text-based pathogen knowledge to both predict antibiotic effectiveness and design novel antibacterials from first principles. Our study yields several key advances.

First, by integrating genomic and textual information, ApexOracle captures a holistic representation of the pathogen that improves predictive accuracy and out-of-distribution generalization. Traditional antimicrobial models depend on expensive isolated strain-specific dataset, whereas ApexOracle infers molecule response directly from a pathogen’s genetic code and documented biological traits. This multimodal strategy is essential for addressing the 'future-pathogen' problem, where no corresponding antimicrobial training data are available. Biology is inherently information-rich: sequence patterns distilled by the genomic language model and knowledge embedded in scientific literature can be recombined to characterize newly encountered organisms. With suitable data, this approach could extend to emergent infectious agents beyond bacteria and fungi—novel viruses, for example.

Second, ApexOracle unifies prediction and generation within a single framework. Earlier AI efforts typically separate (i) predictors that identify which existing drugs might work and (ii) generators that propose new molecules for a fixed pathogen set. Our integrated architecture ensures that every compound it designs is contextualized by a specific pathogen, enabling flexible, pathogen-targeted antibiotic discovery. 

Third, the model’s capacity to explore unconventional chemical space offers a path to outrun resistance by discovering chemotypes bacteria have never encountered and for which no ready defenses exist.

Despite these advances, our work has several limitations. Data scarcity is paramount: ApexOracle’s accuracy is bounded by its training distribution. Pathogens harboring biology or resistance mechanisms absent from our data—e.g., the cell-wall-deficient Mycoplasmatota—may elude correct prediction or yield suboptimal designs. We mitigated this by leveraging Evo2’s broad genomic coverage and diverse textual corpora, yet gaps persist. Moreover, textual embeddings depend on the completeness of scientific knowledge; freshly emerged pathogens often lack detailed annotations, limiting immediate benefit from the text modality. Incorporating real-time data streams, early laboratory observations or rapid screens, could continually refine predictions.

A second limitation concerns objective scope. Generation was optimized primarily for antibacterial potency. ApexOracle currently does not explicitly model toxicity, off-target effects, metabolic stability, or synthetic feasibility. Future work could embed multi-objective optimization—maximizing activity while minimizing predicted toxicity, for instance—through reinforcement learning or Pareto-efficient generative methods.

Translation to the clinic raises practical challenges. Novel molecules may be synthetically complex or costly at scale. Continued collaboration with medicinal chemists is essential to balance novelty against manufacturability. Our use of SELFIES and chemical validity checks provides a baseline safeguard, but coupling the generator with retrosynthesis-planning tools will further streamline route assessment.

In summary, ApexOracle represents a significant step toward next-generation antibiotic discovery. In a world where bacteria continually adapt and new microbes emerge abruptly, the ability to predict and pre-empt threats with targeted therapeutics is invaluable. With continued refinements, including broader data, multi-objective design, and automated synthesis planning, AI platforms like ApexOracle could become integral to our infectious-disease arsenal, scanning genomic data from clinics or environmental samples for resistance genes and instantly proposing effective treatments. This proactive stance—“designing antibiotics of the future for the pathogens of the future”—promises to shorten the lag between pathogen emergence and therapeutic response.

\section{Methods}\label{sec11}
\subsection{Data}

\noindent
\textbf{Antimicrobial Peptides.} 
The antimicrobial peptide (AMP) dataset comprises two components: (1) our in-house peptide collection and (2) the Database of Antimicrobial Activity and Structure of Peptides (DBAASP)~\cite{pirtskhalava2021dbaasp}. From our in-house collection, we obtained 1,642 canonical linear peptides, covering 11 pathogen strains and 15,718 minimum inhibitory concentration (MIC) measurements. From DBAASP, we retrieved 16,408 peptides—71.5\% of which contain noncanonical amino acids or modifications—alongside 5,630 pathogen strains and 105,547 MIC measurements. We then merged the two sources, excluding any strains with incorrect identifiers or names and any peptides containing noncanonical residues that were erroneous or lacked a valid SELFIES representation. After filtering, the consolidated dataset comprises 17,988 peptides, 5,632 pathogen strains, and 121,265 MIC measurements. All peptide sequences were converted to SELFIES strings using \hyperref[peplink]{PepLink}.

All MIC values originally reported in $\mu g/mL$ were converted to $\mu mol$. Values annotated with special operators (e.g., “$\leq$,” “$>$”) were handled according to the procedures summarized in Table \ref{DBAASP_MIC}. For model training, each MIC measurement was further transformed as $-\log_{10}(\frac{\text{MIC}}{10})$.\\

\begin{table}[h]
\renewcommand{\arraystretch}{1.5}  
\caption{DBAASP MIC label pre-processing rules.}\label{DBAASP_MIC}%
\begin{tabular}{@{}p{2cm}l p{2.8cm}@{}}
\toprule
Operators & Transformations & Example\\
\midrule
$>, >=, \geq$    & Double & Record: $> V$\newline Transformed: $2 \times V$\\
$\gg$    & Three times & Record: $\gg V$\newline Transformed: $3 \times V$\\
$ - =>$, $->=$, $ - >=$, $->$, $-$ & Average & Record: $V_a -> V_{b}$\newline Transformed: $\frac{V_a+V_b}{2}$\\
$\pm$    & Average & Record: $V_a \pm V_{b}$\newline Transformed: $V_a$\\
Others    & Keep & Record: $V_a$\newline Transformed: $V_a$\\
\botrule
\end{tabular}
\end{table}

\noindent
\textbf{PepLink.}\label{peplink} We developed PepLink, a versatile converter that transforms amino acid sequences—including noncanonical residues, intrachain bonds, and terminal modifications—into SELFIES strings, and also supports reverse conversion from SELFIES back to amino acid sequences. It currently supports 404 distinct noncanonical amino acids, 11 intrachain-bond types, 242 N-terminal modifications, and 56 C-terminal modifications. The conversion proceeds in three steps: (1) concatenating the SMILES of canonical and noncanonical residues into a linear backbone; (2) introducing specified intrachain bonds to generate cyclic or bridged structures; and (3) appending the desired N- and C-terminal modifications.

All noncanonical residues were curated from DBAASP. We retrieved SMILES for the 20 canonical amino acids—and for many noncanonical ones—directly from PubChem. For those DBAASP entries lacking PubChem records, we used ChatGPT-o1 to standardize their DBAASP-provided IUPAC names, then converted these names to SMILES via OPSIN \cite{lowe2011chemical}. In total, this process yielded 404 unique noncanonical amino acids present in the DBAASP dataset. See Appendix.\ref{secB} for details.
\\

\noindent
\textbf{Small Molecule Antibiotics.} The small molecule antibiotic dataset comprises 49,331 \textit{(molecule, strain)} pairs with binary classification labels of antibiotic activity against three different bacterial strains: \textit{S. aureus} RN4220 (39,312 molecules from~\cite{wong2024discovery}), \textit{E. coli} BW25113 (2,335 molecules from~\cite{stokes2020deep}), and \textit{A. baumannii} ATCC 17978 (7,684 molecules from~\cite{liu2023deep}). 
\\

\noindent
\textbf{Genome Data.} Pathogen genome data were downloaded from the ATCC Genome Portal, except for the following four strains: \textit{E. coli} MG1665, \textit{E. coli} UMNK88, \textit{P. aeruginosa} PA14, and \textit{S. aureus} USA300. Since the genome information of these four strains is not recorded in ATCC, we instead obtained their data from NCBI. See Appendix.\ref{secC} for details..
\\

\noindent
\textbf{Strain Description Data.} Leveraging the powerful search and memory capabilities of modern large language models, we utilized Qwen2.5-Max~\cite{qwen25} (version 01.25.2025) to search for important strain traits from reliable scientific publications. These traits include:
\begin{enumerate}[1.]
\item Species Information: Description of the species to which this strain belongs, including whether it is Gram-positive, Gram-negative, Fungi, Archaea, or Protozoa. Also, its notable physiological traits.

\item Unique Mutations: Any distinctive genetic mutations recorded to have impact on metabolic pathways or plasma membrane. And how these mutations modify bacterium's behavior and develop antimicrobial resistance.

\item Antibiotics or antimicrobial peptides resistance and sensitivity: Antibiotics and antimicrobial peptides to which this strain is known to be sensitive or resistant.
\end{enumerate}
See Appendix.\ref{secA1} for the prompt.
\\

\noindent
\textbf{Molecules for diffusion language model training.} We curated a diverse set of sequences from five sources: PubChem (downloaded via MolFormer \cite{ross2022large}; 111,378,206 molecules), peptides from SmProt v2.0 (825,632 sequences), UniRef ($<=$50 residues; 6,972,866 sequences), UniProt ($<=$50 residues; 3,749,540 sequences), and modification-rich peptides generated by CycloPS from PeptideCLM \cite{peptideclm} (10,000,000 sequences). 
We used IBM's SELFIES \hyperlink{https://huggingface.co/ibm-research/materials.selfies-ted}{tokenizer}~\cite{ross2022large} to parse these molecules. Any molecule having more than 1,024 tokens after tokenization was discarded, and duplicate entries between UniRef and UniProt were removed. After these filtering steps, we obtained 121.6 million molecules. We further used RDKit to calculate  a 209-dimensional property vector for these molecules (Fig.\hyperref[fig:all]{1b}), which quantifies a molecule’s size, topology, electronic and surface properties, and functional group composition. We randomly withheld 1\% of the data for evaluation and used the remaining 99\% for training.
\\

\noindent
\textbf{Synergy Data.} All synergy data were curated from the DBAASP database, encompassed a broader range of compounds than solely AMPs. In total, our curated dataset comprised 2,732 unique molecule synergy-strain pairs, of which 88\% were AMP-small molecule interactions and the remaining 12\% were AMP-AMP interactions. Recognizing the challenges posed by the scarcity and inherent noise in synergy data, we binarized the FICI values using a 0.5 cutoff—labeling values below 0.5 as synergistic (1) and those equal to or above 0.5 as indifferent (0).

\subsection{ApexOracle Architecture and Training}

\noindent
\textbf{Genome Embeddings.} Pathogen genomes downloaded from ATCC typically consist of multiple contigs. For each contig, we employ a sliding window approach to fragment it. Specifically, we use a step size of 10,000 nt and a window length of 11,000 nt, ensuring that each window is 1,000 bases longer than the step size to prevent genes from being truncated and thus feed imperfect information to the downstream model. Each extracted fragment is then inputted into Evo-2 (40B version), from which we extract the latent embedding of the $46^{th}$ layer for all bases. We compute the contig-level embedding by averaging its per-base embeddings. By stacking the embeddings of all fragments' representations, we generate the embedding of the entire genome. 

Suppose the genome yields $M$ fragments and let $\mathbf{e}_i\in\mathbb{R}^{d_g}$ denote the embedding of fragment $i$ (with length $L_i$), computed as
\[
\mathbf{e}_i \;=\; \frac{1}{L_i}\sum_{t=1}^{L_i} \mathbf{h}_{i,t}^{(46)}.
\]
where $\mathbf{h}_{i,t}^{(46)}\in\mathbb{R}^{d_g}$ is the hidden state vector of the $t$-th nucleotide in fragment $i$ as produced by the $46^{th}$ transformer layer of Evo-2. Then the full-genome embedding is given as
\[
\mathbf{E} = [\mathbf{e}_1,\, \mathbf{e}_2,\, \dots,\, \mathbf{e}_M] \;\in\; \mathbb{R}^{M\times d_g}.
\]
We found that the embeddings produced by Evo2-40B exhibit magnitudes concentrated on the order of $10^{-15}$, whereas the ApexOracle model’s weights are initialized around  $10^{-1}$, to reconcile this scale mismatch and avoid numerical issues, we therefore uniformly scale all genome embeddings by a factor of $10^{14}$.
\\

\noindent
\textbf{Text Embeddings.} For pathogen strain description obtained from Qwen2.5-Max~\cite{qwen25} (version 01.25.2025).  We utilize Me-LLaMA~\cite{Me-llama} to obtain embeddings for pathogen strain description texts. Specifically, we employ the Me-LLaMA3-8B model, which has been further pretrained on large-scale open-source medical data, to generate these embeddings. For each strain, the descriptive text is first preprocessed by replacing the specific strain name with “this strain” to prevent overfitting to particular strain names. The text is then input into Me-LLaMA, and we extract the latent embedding from the penultimate layer as the text representation. Unlike the genome embeddings obtained from Evo-2, we do not apply mean pooling to these text embeddings, allowing ApexOracle to capture more precise semantic information.

Suppose the preprocessed text is tokenized into \(N\) tokens and let \(\mathbf{t}_j^{(p)}\in\mathbb{R}^{d_t}\) denote the hidden state vector of the \(j\)-th token from the penultimate layer of Me-LLaMA3-8B. Then the full text embedding for the strain is
\[
\mathbf{T} = [\mathbf{t}_1^{(p)},\, \mathbf{t}_2^{(p)},\, \dots,\, \mathbf{t}_N^{(p)}] \;\in\; \mathbb{R}^{N\times d_t}.
\]
\\

\noindent
\textbf{Diffusion language model.}\label{MolMDLM} Diffusion models are latent variable generative models characterized by a forward and a reverse  Markov process. The forward process $q(\boldsymbol{x}_{1:T}|\boldsymbol{x}_0) = \prod_{t=1}^T q(\boldsymbol{x}_t|\boldsymbol{x}_{t-1})$ corrupts the data $\boldsymbol{x}_0 \sim q(\boldsymbol{x}_0)$ into a sequence of increasingly noisy latent variables $\boldsymbol{x}_{1:T} = \boldsymbol{x}_1, \boldsymbol{x}_2, ..., \boldsymbol{x}_T$. The learned reverse Markov process $p_{\theta}(\boldsymbol{x}_{0:T}) = p(\boldsymbol{x}_{T})\prod_{t=1}^T p_{\theta}(\boldsymbol{x}_{t-1}|\boldsymbol{x}_{t})$ gradually denoises the latent variables towards the data distribution, where $\theta$ represents the learnable parameters.

In our DLM, both the forward and reverse processes operate in the discrete SELFIES token space. During the forward process, noise is progressively added by replacing actual SELFIES tokens with \texttt{<MASK>} tokens. In the reverse process, a language model reconstructs the original SELFIES tokens from these \texttt{<MASK>} tokens. Let $\mathbf{x}_0$ be a $K$-dimensional one-hot encoding of a SELFIES token, where $K$ stands for the vocabulary size of SELFIES token space (note that special tokens like \texttt{<MASK>} are included in the space). We followed the Masked Diffusion Language Model (MDLM) \cite{sahoo2024simple} workflow to define the absorbing-state forward diffusion process $q(\mathbf{x}_t | \mathbf{x}_0)$ as 
\begin{equation}
q(\mathbf{x}_t | \mathbf{x}_0) = \text{Cat}(\mathbf{x}_t; \alpha_t \mathbf{x}_0 + (1 - \alpha_t) \mathbf{m}).
\end{equation}
Here, $\alpha_t \in [0, 1]$ is a strictly decreasing function with respect to $t$, where $\alpha_0 \approx 1$ and $\alpha_1 \approx 0$, representing the probability that $\mathbf{x}_0$ remains unchanged at time $t \in [0,1]$. The vector $\mathbf{m}$ denotes the one-hot encoding of the \texttt{<MASK>} token, and $\text{Cat}(\cdot; \boldsymbol{\pi})$ denotes a categorical distribution over the $K$ SELFIES tokens, parameterized by the probability simplex $\boldsymbol{\pi}$. The posterior $q(\mathbf{x}_{t-1} | \mathbf{x}_t, \mathbf{x}_0)$ can be written as

\begin{equation}
\begin{split}
&q\bigl(\mathbf{x}_{t-1} \mid \mathbf{x}_t, \mathbf{x}_0\bigr) = \\
&\begin{cases}
\mathrm{Cat}\bigl(\mathbf{x}_{t-1};\, \mathbf{x}_t\bigr), 
& \mathbf{x}_t \neq \mathbf{m}, \\[1ex]
\mathrm{Cat}\!\Bigl(\mathbf{x}_{t-1};\,
\frac{(1 - a_{t-1})\,\mathbf{m} + (a_{t-1} - a_t)\,\mathbf{x}_0}{1 - a_t}\Bigr), 
& \mathbf{x}_t = \mathbf{m}.
\end{cases}
\end{split}
\tag{2}\label{equ2}
\end{equation}

In practice, we don't have access to the ground truth $\mathbf{x}_0$ in $q(\mathbf{x}_{t-1} | \mathbf{x}_t, \mathbf{x}_0)$. Therefore, we seek to use a neural network $\text{NN}_{\theta}(\cdot, \cdot)$ parameterized by $\theta$ to take the time step $t$ and its corresponding noisy sample $\mathbf{x}_t$ as inputs and estimate the clean data $\mathbf{x}_0$:

\begin{equation}
\begin{split}
&p_\theta\bigl(\mathbf{x}_{t-1} \mid \mathbf{x}_t, t\bigr) = \\
&\begin{cases}
\mathrm{Cat}\bigl(\mathbf{x}_{t-1};\, \mathbf{x}_t\bigr), 
& \mathbf{x}_t \neq \mathbf{m}, \\[1ex]
\mathrm{Cat}\!\Bigl(\mathbf{x}_{t-1};\,
\frac{(1 - a_{t-1})\,\mathbf{m} + (a_{t-1} - a_t)\,\text{NN}_{\theta}(\mathbf{x}_t,t) }{1 - a_t}\Bigr), 
& \mathbf{x}_t = \mathbf{m}.
\end{cases}
\end{split}
\tag{3}\label{equ3}
\end{equation}
Regarding the output $\text{NN}_{\theta}(\mathbf{x}_t,t)$, we follow the SUBS parameterization in MDLM to further replace the logit output corresponding to the \texttt{<MASK>} token by $-\infty$ to make sure the neural network denoiser does not output \texttt{<MASK>} tokens.

Assuming we have a sequence of $L$ SELFIES tokens $\mathbf{x}_t^{1:L}$ at time $t$, and that the tokens in $\mathbf{x}_{t-1}^{1:L}$ are conditionally independent given $\text{NN}_{\theta}(\mathbf{x}_t^{1:L}, t)$, then under a log-linear noise schedule, where $\alpha_t = \exp(-r(t))$ and $r(t) = -\log(1 - t)$, the optimization objective of the DLM—namely, the negative evidence lower bound—can be defined as:

\begin{align}
\mathcal{L}_{\text{DLM}} 
&= -\mathbb{E}_{t \sim \mathcal{U}(0,1],\, q(\mathbf{x}_t^{1:L} \mid \mathbf{x}_0^{1:L})} \notag \\
&\quad \left[
\frac{1}{t}
\sum_{l=1}^{L} \log \left\langle \text{NN}_\theta^l (\mathbf{x}_t^{1:L}, t),\, \mathbf{x}_0^l \right\rangle
\right].
\tag{4}
\end{align}
Here $\mathcal{U}(0,1]$ is an uniform distribution, and $<\cdot,\cdot>$ stands for dot product. We implemented our $\text{NN}_{\theta}(\cdot, \cdot)$ as a 12-layer Diffusion Transformer (DiT)~\cite{DiT} with a latent dimension of 768.

As $\mathcal{L}_{\text{DLM}}$ can be viewed as a weighted average of masked language modeling losses over different noise levels, we think DLM can also be used as a feature extractor from clean data inputs. Inspired by~\cite{ahmad2022chemberta}, we further enhance the representation learning capability of our DLM by introducing an additional multilayer perceptron (MLP), which takes as input the embedding of the beginner token (i.e., the \texttt{<CLS>} token) from the final transformer layer of DiT, computed from a clean SELFIES string. The MLP is trained to predict the corresponding 209-dimensional vectorized computational descriptors generated by RDKit for the same SELFIES string. We calculate the mean squared loss between predictions and labels, and term this loss as multitask regression loss (MTR) $\mathcal{L}_{\text{MTR}}$. Our final loss to train the DML is a weigthed combination of the two losses defined above:
\begin{equation}
\mathcal{L}_{\text{ApexOracle-DLM}}=\mathcal{L}_{\text{DLM}}+\lambda\cdot\mathcal{L}_{\text{MTR}},
\tag{5}
\end{equation}
where we simply set $\lambda=0.1$. 
\\

\noindent
\textbf{Molecule-Strain Knowledge Fusion.} Once we have a trained molecule DLM, we can feed a clean molecule SEIFIES string into DLM and treat the beginner token embedding from the last transformer layer of DiT as the its feature representation. ApexOracle then uses MLPs to transform molecule feature representations, the genome and text embeddings of pathogens into new hidden spaces, and fuse the molecular information with pathogen strain-information via the cross attention mechanism. Here we treat the molecule’s representation as a single query Q that attends in parallel to two separate pathogen strain-knowledge banks: one comprising genome-derived key–value pairs $(K_{\rm genome},V_{\rm genome})$ and the other comprising text-derived key–value pairs $(K_{\rm text},V_{\rm text})$ (Fig.\hyperref[fig:all]{1a}). In each cross-attention block, the molecule query ($Q$) computes attention weights over the respective keys and then updates itself with the corresponding values via a residual connection. The resulting molecule-genome and molecule-text context vectors are concatenated into a unified embedding that seamlessly combines molecular features with pathogen strain-specific insights. This fused representation is finally passed through task-specific MLP heads to predict antibiotic classification, minimum inhibitory concentration (MIC), and fractional inhibitory concentration index (FICI).
\\

\noindent
\textbf{Task MLPs.}\label{MLP} ApexOracle employs three task-specific MLP heads (Fig.\hyperref[fig:all]{1a}): one for antibiotic classification, one for MIC regression, and one for synergy prediction. The antibiotic classification and MIC regression heads share the same architecture (layer size: 12,294$\rightarrow$3,073$\rightarrow$128$\rightarrow$1); the former is trained with binary cross-entropy loss on small molecule antibiotics datasets as these datasets offer only binary labels, and the latter with mean-squared error loss on our merged AMPs dataset. For synergy prediction, each molecule is processed through the fusion module in parallel to produce two unified embeddings, which are then concatenated and fed into the synergy MLP head. To accommodate the combined input, the first layer of the synergy head has twice the input dimension (layer size: 12,294$\times2$$\rightarrow$3,073$\rightarrow$128$\rightarrow$1), while all other architectural parameters remain identical. The synergy head itself is trained using binary cross-entropy loss. 
\\

\noindent
\textbf{Predictor-Based Guidance.} 
Generating molecules with desired properties is crucial, as random generation followed by predictor screening is highly inefficient and often fails to produce molecules with all the required characteristics simultaneously. Here, we employed discrete predictor-based guidance ~\cite{cbg} to steer ApexOracle towards generating molecules with desired antimicrobial properties. Without loss of generality, we consider a single property predictor guided scenario to elucidate how the guided generation works and extend it to multi-predictor guided cases. Let the predictor be the antibiotic classification predictor $p(y |\mathbf{x}_t^{1:L}, t) \in [0, 1]$ from our \hyperref[MLP]{Task MLPs} and is trained on noisy data $\mathbf{x}_t^{1:L}$, where $y$ is the binary antimicrobial label. Noisy predictors are trained to address the limitation that models trained only on clean data often provide poor guidance for noisy inputs. During guided generation, we utilize these noisy predictors, while reserving the clean-data-trained predictors solely for reporting the final properties of the generated molecules. Following Bayes' rule, we have $p_{\theta}(\mathbf{x}_{t-1}^{1:L} | \mathbf{x}_{t}^{1:L}, t, y) \propto p(y | \mathbf{x}_{t-1}^{1:L}, \mathbf{x}_{t}^{1:L}, t) p_{\theta}(\mathbf{x}_{t-1}^{1:L} |\mathbf{x}_{t}^{1:L}, t)$. Under the assumption that tokens in $\mathbf{x}_{t-1}^{1:L}$ are conditionally independent given $\mathbf{x}_{t}^{1:L}$, we have the token level proportion:
$p_{\theta}(\mathbf{x}_{t-1}^{l} | \mathbf{x}_{t}^{1:L}, t, y) \propto p(y | \mathbf{x}_{t-1}^{1:L}, \mathbf{x}_{t}^{1:L}, t) p_{\theta}(\mathbf{x}_{t-1}^{l} |\mathbf{x}_{t}^{1:L}, t)$. Here, $p_{\theta}(\mathbf{x}_{t-1}^{l} |\mathbf{x}_{t}^{1:L}, t)$ can be directly obtained from the DLM while $p(y | \mathbf{x}_{t-1}^{1:L}, \mathbf{x}_{t}^{1:L}, t)$ requires further approximation. We follow the practice from~\cite{cbg} to define $\hat{\mathbf{x}}_{t, l}^{1:L}$ as 
$[\mathbf{x}_t^{1,l-1}, \mathbf{x}_{t-1}^l, \mathbf{x}_t^{1+1:L}]$. That is, $\hat{\mathbf{x}}_{t, l}^{1:L}$ is an $L$-token sequence in which the $l$-th token is taken from $\mathbf{x}_{t-1}^l$, while all other tokens are the same as those in $\mathbf{x}_{t}^{1:L}$. We then use $p(y | \hat{\mathbf{x}}_{t, l}^{1:L}, t)$ to approximate $p(y | \mathbf{x}_{t-1}^{1:L}, \mathbf{x}_{t}^{1:L}, t)$. Now, we can write the predictor-guided generation probability for a SEIFIES string as
\begin{small}
\begin{equation}
\begin{split}
&p_{\theta}\bigl(\mathbf{x}_{t-1}^{(1:L)} \mid \mathbf{x}_t^{(1:L)},\,t,\,y\bigr)
= \prod_{\ell=1}^L \\[1ex]
&\frac{
    p(y | \hat{\mathbf{x}}_{t, l}^{1:L}, t)\,
    p_{\theta}\bigl(\mathbf{x}_{t-1}^{(\ell)} \mid \mathbf{x}_t^{(1:L)}, t\big)
}{
    \displaystyle
    \sum_{\hat{\mathbf{z}}_{t,l}^{1:L}}
    p(y | \hat{\mathbf{z}}_{t, l}^{1:L}, t)\,
    p_{\theta}\bigl(\mathbf{x}_{t-1}^{(\ell)} \mid \mathbf{x}_t^{(1:L)}, t\big)
},
\end{split}
\tag{6}\label{equ6}
\end{equation}
\end{small}
where $\hat{\mathbf{z}}_{t, l}^{1:L}$ is an $L$-token sequence $[\mathbf{x}_t^{1,l-1}, \mathbf{z}, \mathbf{x}_t^{1+1:L}]$ in which the $l$-th token can be the one-hot encoding of any token in the SEIFIES vocabulary. By summing over all possible tokens at the $l$-th position in the SELFIES string, the denominator in Equation (\ref{equ6}) normalizes the expression, ensuring it represents a valid probability distribution.

To extend this guided generation framework to multiple property predictors, let us, without loss of generality, consider two properties, $y_1$ and $y_2$. By assuming that $y_1$ and $y_2$ are conditionally independent given $\mathbf{x}_t^{1:L}$, we have $p(y_1, y_2 | \mathbf{x}_{t}^{1:L}, t)$ = $p(y_1 | \mathbf{x}_{t}^{1:L}, t) p(y_2 | \mathbf{x}_{t}^{1:L}, t)$. By further introducing two extra hyperparameters $\gamma_1$ and $\gamma_2$ to control the guidance strengths of predictors, the new predictor-guided generation probability can now be rewritten as:

\begin{small}
\begin{equation}
\begin{split}
&p_{\theta}^{\gamma_1, \gamma_2}\bigl(\mathbf{x}_{t-1}^{(1:L)} \mid \mathbf{x}_t^{(1:L)},\,t,\,y_1, y_2\bigr)
= \prod_{\ell=1}^L \\[1ex]
&\frac{
    p(y_1 | \hat{\mathbf{x}}_{t, l}^{1:L}, t)^{\gamma_1}\,
    p(y_2 | \hat{\mathbf{x}}_{t, l}^{1:L}, t)^{\gamma_2}\,
    p_{\theta}\bigl(\mathbf{x}_{t-1}^{(\ell)} \mid \mathbf{x}_t^{(1:L)}, t\big)
}{
    \displaystyle
    \sum_{\hat{\mathbf{z}}_{t,l}^{1:L}}
    p(y_1 | \hat{\mathbf{z}}_{t, l}^{1:L}, t)^{\gamma_1}\,
    p(y_2 | \hat{\mathbf{z}}_{t, l}^{1:L}, t)^{\gamma_2}\,
    p_{\theta}\bigl(\mathbf{x}_{t-1}^{(\ell)} \mid \mathbf{x}_t^{(1:L)}, t\big)
}.
\end{split}
\tag{7}\label{equ7}
\end{equation}
\end{small}
Suppose the label $y_1$ is a continuous value and the corresponding regressor with parameter $\Phi_1$ noted as $f_{\Phi_1}(\mathbf{x}_{t}^{1:L}, t)$ gives the prediction $y_1'$, we can then use the following transformation  
\begin{equation}
\begin{split}
&E[y_1, y_1'] =  \\ 
&2 \cdot \int_{-\infty}^{\,y_1 - \lvert y_1 - y_1' \rvert} 
\frac{1}{\sqrt{2\pi\,\sigma^2}}
\exp\!\Bigl(-\frac{(y_1 - y_1')^2}{2\,\sigma^2}\Bigr)\,\mathrm{d}t,
\end{split}
\tag{8}
\end{equation}
to replace $p(y_1 | \mathbf{x}_{t}^{1:L}, t)$ so that the generated molecules will be guided to have $y_1'$ close to the target value. 

\noindent
\textbf{Remask during generation.} Denoising process as in Equation (\ref{equ3}) and Equation (\ref{equ7}) can lead to the drawback that once a token is generated, it cannot be updated again, even when it introduced an error. Remasking sampler \cite{remask} can ease this problem by giving the generated tokens a chance to become \texttt{<MASK>} again and be denoised again. Suppose at time $t$, an unmasked token $\mathbf{x}_{t}$ has $r_t$ probability to be remasked, then we can modify the denosing step described in Equation (\ref{equ3}) to:
\begin{equation}
\begin{split}
&p_{\theta}\bigl(\mathbf{x}_{t-1} \mid \mathbf{x}_t, t\bigr) = \\
&\begin{cases}
\mathrm{Cat}\bigl(\mathbf{x}_{t-1};\, (1-r_t)\mathbf{x}_t+r_t\mathbf{m}\bigr),
& \mathbf{x}_t \neq \mathbf{m} \\[1ex]
\mathrm{Cat}\!\Bigl(\mathbf{x}_{t-1};\,
\frac{\beta_1\,\mathbf{m} + \beta_2\,\text{NN}_{\theta}(\mathbf{x}_t, t)}{1 - a_t}\Bigr),
& \mathbf{x}_t = \mathbf{m},
\end{cases} \\
\end{split}
\tag{9}\label{equ9}
\end{equation}
where $\beta_1 = (1 - \alpha_{t-1}-r_t\alpha_t)$, and $\beta_2 = (\alpha_{t-1} - (1-r_t)\alpha_t)$. The detailed remasking schedule $r_t$ will be described in the Sampling Strategy section.
\\

\noindent
\textbf{Sampling Strategy.} We adopted the \textit{Loop} sampling strategy in \cite{remask} and paired it with a three-stage guidance to make sure the generated molecules have low MIC against the target strain and are more likely to be peptides. The MIC regressor from the ApexOrcale as well as a peptide classifier trained on our DLM training data were used as guided predictors. Let $\gamma_1$ and $\gamma_2$ denote the guidance strength for the MIC regressor and peptide classifier, respectively.

It is believed that there may be certain periods of the generation process where remasking is more valuable. Therefore we choose to turn on remasking between timestep $t_{on}, t_{off}\in(0, 1)$. Furthermore, in the range when remasking is activated, we modify the noise schedule to be constant, such that $\alpha_t = \alpha(t_{on})$. In our case, in the first stage, from $t=1$ to $t=t_{on}$, the model generates tokens without remasking (i.e., $r_t = 0$) and is only guided by the MIC regressor ($\gamma_2=0$). In the second stage, from $t=t_{on}$ to $t=t_{off}$, we hold $\alpha$ constant (i.e., $\alpha_{t-1}=\alpha_t$), and the model can ‘correct potential mistakes’ by remasking and predicting a fixed proportion of the generated tokens in a loop. Additionally, the model is only guided by the peptide classifier ($\gamma_1=0$) in the second stage to correct the potential mistakes that lead to generating a molecule that's not a peptide. Finally, the third stage , from $t=t_{off}$ to $t=0$, is the same as stage one, we let the model predict any remaining unmasked tokens using the DLM posterior.
\\

\noindent
\textbf{Implementation Details.} We tokenized SELFIES strings using MolFormer’s SELFIES tokenizer \cite{ross2022large} and retained 121.6 million sequences of up to 1,024 tokens, each paired with a multi-target vector. We randomly withheld 1\% of the data for evaluation and used the remaining 99\% for training. Training proceeded in two stages: an initial stage with one training epoch to optimize only the DLM objective, followed by a second stage with two training epochs to incorporate the MTR objective into training. For the MTR objective, we utilized RDKit to generate 209 descriptors of each molecule and used the model to predict the 209 descriptors. To extract features from the clean data, we set $t=\epsilon$, where $\epsilon=10^{-3}$ in our case, indicating the last step of generation with ignorable noise. Setting $t=0$ gets nearly the same performance while setting $t=\epsilon$ is more compatible with the generation noise schedule. A peak learning rate of $3\times10^{-4}$  and a batch size of 768 were used in the first, with a learning rate linear warm-up over the first 2,500 steps. In the second stage, the learning rate was held constant at $10^{-4}$ and the batch size was set to 480. Optimization employed AdamW on eight NVIDIA A100 GPUs (80 GB each). After both stages, the model achieved a perplexity of 1.58 and an NLL of 0.46 on the validation set.

For synergy prediction, we first trained ApexOracle for 13 epochs using the complete MIC and antibiotic classification datasets, and saved its parameters as a base model for subsequent fine-tuning on the synergy task. We then added a new synergy MLP head and fine-tuned the molecule-strain knowledge fusion module using LoRA with a rank of 64. We ensembled 7 models to get the reported result.

For guided generation, we set a total of 256 steps for the model to generate. We set target MIC as 1 to guide the model toward generating molecules with low MIC values. The value of $\sigma$ in $p(y_1 | \mathbf{x}_{t}^{1:L}, t)$ was linearly annealed from 0.5 to 0.2 as time progressed from 1 to 0. We set $t_{on}=0.55, t_{off}=0.45$, giving the model a chance to ‘correct mistakes’ in the middle of the generation process. During phase one and three, we set $\gamma_1=15, \gamma_2=0$. During phase two, we set $\gamma_1=0, \gamma_2=15$. Both the MIC regressor and the peptide classifier used for guided generation are trained with noised sequences.

\section*{Acknowledgements}

Cesar de la Fuente-Nunez holds a Presidential Professorship at the University of Pennsylvania and acknowledges funding from the Procter \& Gamble Company, United Therapeutics, a BBRF Young Investigator Grant, the Nemirovsky Prize, Penn Health-Tech Accelerator Award, Defense Threat Reduction Agency grants HDTRA11810041 and HDTRA1-23-1-0001, the Langer Prize (AIChE Foundation), and the Dean’s Innovation Fund from the Perelman School of Medicine at the University of Pennsylvania. Research reported in this publication was supported by the NIH R35GM138201 and DTRA HDTRA1-21-1-0014. 

\section*{Declarations}

\begin{itemize}

\item Data availability 

The data used for our DLM pre-training can be found at \hyperlink{https://huggingface.co/datasets/Kiria-Nozan/ApexOracle}{https://huggingface.co/datasets/Kiria-Nozan/ApexOracle}
\item Code availability 

The code, DLM weights, and necessary materials used in the study can be found at \hyperlink{https://github.com/DragonDescentZerotsu/ApexOracle}{https://github.com/DragonDescentZerotsu/ApexOracle}, \hyperlink{https://huggingface.co/Kiria-Nozan/ApexOracle}{https://huggingface.co/Kiria-Nozan/ApexOracle}

and \hyperlink{https://zenodo.org/records/15612048}{https://zenodo.org/records/15612048}
\item Author contribution

T.L., F.W., C.F.N. conceived the idea. C.F.N. acquired funding, designed, administered, and supervised this work; 
T.L. build the ApexOracle architecture with F.W.'s advice; F.W. trained the DLM model; 
T.L., F.W. and C.F.N. wrote the initial version of the manuscript. All co-authors edited subsequent versions of the manuscript.

\end{itemize}

\bibliography{sn-bibliography}

\begin{appendices}

\lstset{
  basicstyle=\ttfamily\scriptsize,   
  breaklines=true,                    
  breakatwhitespace=true,             
  numbers=none,                       
  numberstyle=\tiny\color{gray},      
  frame=none,                       
  xleftmargin=0pt, xrightmargin=0pt,  
  keywordstyle=\color{blue},          
  commentstyle=\color{gray},          
  stringstyle=\color{red},            
  showstringspaces=false              
}

\section{Non-canonical Amino Acids}\label{secB}
\begin{lstlisting}[breaklines=true, breakatwhitespace=false, columns=fullflexible]
a: C[C@H](C(=O)O)N
r: C(C[C@H](C(=O)O)N)CN=C(N)N
n: C([C@H](C(=O)O)N)C(=O)N
d: C([C@H](C(=O)O)N)C(=O)O
c: C([C@H](C(=O)O)N)S
q: C(CC(=O)N)[C@H](C(=O)O)N
e: C(CC(=O)O)[C@H](C(=O)O)N
h: C1=C(NC=N1)C[C@H](C(=O)O)N
i: CC[C@@H](C)[C@H](C(=O)O)N
l: CC(C)C[C@H](C(=O)O)N
k: C(CCN)C[C@H](C(=O)O)N
m: CSCC[C@H](C(=O)O)N
f: C1=CC=C(C=C1)C[C@H](C(=O)O)N
P: C1C[C@H](NC1)C(=O)O
s: C([C@H](C(=O)O)N)O
t: C[C@@H]([C@H](C(=O)O)N)O
w: C1=CC=C2C(=C1)C(=CN2)C[C@H](C(=O)O)N
y: C1=CC(=CC=C1C[C@H](C(=O)O)N)O
v: CC(C)[C@H](C(=O)O)N
1-NAL: C[C@@](C1=CC=CC2=CC=CC=C21)(C(=O)O)N
7-NH2-C7: C(CCCN)CCC(=O)O
12-NH2-C12: C(CCCCCC(=O)O)CCCCCN
TYR-Bzl: C1=CC=C(C=C1)COC2=CC=C(C=C2)C[C@@H](C(=O)O)N
D-ORN: C(C[C@H](C(=O)O)N)CN
NLE: CCCC[C@@H](C(=O)O)N
ORN: C(C[C@@H](C(=O)O)N)CN
AIB: CC(C)(C(=O)O)N
ABU: CCC(C(=O)[O-])N
D-NLE: CCCC[C@H](C(=O)O)N
P-Ser: C(C(C(=O)O)N)OP(=O)(O)O
PYA: C1CC(=O)N[C@@H]1C(=O)O
ADOO: C(COCCOCC(=O)O)N
DHB: CC=C(C(=O)O)N
P-Tyr: C1=CC(=CC=C1C[C@@H](C(=O)O)N)OP(=O)(O)O
MET-OXD: CS(=O)CCC(C(=O)O)N
D-Allo-ILE: CC[C@H](C)[C@H](C(=O)O)N
bALA: C(CN)C(=O)O
D-PYA: C1CC(=O)N[C@H]1C(=O)O
Et-GLU: CCOC(=O)CC[C@@H](C(=O)O)N
5-Me-GLU: COC(=O)CC[C@@H](C(=O)O)N
CIT: C(C[C@@H](C(=O)O)N)CNC(=O)N
NVAL: CC(C)NCC(=O)O
Sar: CNCC(=O)O
4-HYP: C1C(CNC1C(=O)O)O
Me-Leu: CC(C)CC(C(=O)O)NC
Cha: C1CCC(CC1)CC(C(=O)O)N
3-Me-Val: CC(C)(C)C(C(=O)O)N
Ac6c: C1CCC(CC1)(C(=O)O)N
bhARG: C(C[C@@H](CC(=O)O)N)CN=C(N)N
HARG: C(CC(C(=O)O)N)CN=C(N)N
NOR: C(C[C@@H](C(=O)O)N)CN=C(N)N[N+](=O)[O-]
DOPA: C1=CC(=C(C=C1C[C@@H](C(=O)O)N)O)O
CCA: C1=CC=C(C(=C1)C(=O)[O-])NC2=C(C=CC(=C2)Cl)C(=O)[O-].[Na+].[Na+]
6-Br-trp: C1=CC2=C(C=C1Br)NC=C2CC(C(=O)O)N
AGL: N[C@@H](CC=C)C(=O)O
LYS-(CH3)3: C[N+](C)(C)CCCC[C@@H](C(=O)[O-])[NH3+]
AdaGly: C1C2CC3CC1CC(C2)(C3)NCC(=O)O
Ada-Ala: C1C2CC3CC1CC(C2)(C3)CC(C(=O)O)N
Chg: C1CCC(CC1)C(C(=O)O)N
3,4-OH-ARG: C(C(C([C@@H](C(=O)O)N)O)O)N=C(N)N
4,5-OH-LYS: OC(C[C@H](N)C(=O)O)C(CN)O
5-OH-LYS: C(C[C@@H](C(=O)O)N)[C@H](CN)O
D-HPhe: C1=CC=C(C=C1)CC[C@H](C(=O)O)N
D-1-NAL: C1=CC=C2C(=C1)C=CC=C2C[C@H](C(=O)O)N
D-2-NAL: C1=CC=C2C=C(C=CC2=C1)C[C@H](C(=O)O)N
D-DIP: C1=CC=C(C=C1)C(C2=CC=CC=C2)[C@H](C(=O)O)N
D-TIC: C1[C@@H](NCC2=CC=CC=C21)C(=O)O
APC: C1CC(N(C1)C(=O)O)N
Ac5c: C1CCC(C1)(C(=O)O)N
DAP: C(C(C(=O)O)N)N
DHF: NC(=CC1=CC=CC=C1)C(=O)O
MET-O2: CS(=O)(=O)CC[C@@H](C(=O)O)N
5-Br-TRP: C1=CC2=C(C=C1Br)C(=CN2)CC(C(=O)O)N
D-Allo-Thr: C[C@H]([C@H](C(=O)O)N)O
D-NVA: CCC[C@H](C(=O)O)N
NArg: N(C(=N)N)CCCNCC(=O)O
Ahx: C(CCC(=O)O)CCN
N,N-Me-ILE: CC[C@H](C)[C@@H](C(=O)O)N(C)C
GABA: C(CC(=O)O)CN
4-F-PHE: C1=CC(=CC=C1C[C@@H](C(=O)O)N)F
4-NO2-PHE: C1=CC(=CC=C1CC(C(=O)O)N)[N+](=O)[O-]
Allo-ILE: CC[C@@H](C)[C@@H](C(=O)O)N
TOAC: CC1(CC(CC(N1[O])(C)C)(C(=O)O)N)C
Me-Ser: CN[C@@H](CO)C(=O)O
2-NAL: C1=CC=C2C=C(C=CC2=C1)C[C@@H](C(=O)O)N
2-NH2-C11: CCCCCCCCC[C@@H](C(=O)O)N
DHV: NC(C(=O)O)=C(C)C
DEtGly: CCN(CC)CC(=O)O
Ac3c: NC1(CC1)C(=O)O
Ac4c: NC1(CCC1)C(=O)O
Aic: C1=NC(=C(N1)C(=O)N)N
ACPC: C1CC1(C(=O)O)N
LEUol: CC(C)C[C@@H](CO)N
ADMArg: CN(C)C(=NCCC[C@@H](C(=O)O)N)N
SDMArg: CNC(=NC)NCCC[C@@H](C(=O)O)N
PHEol: C1=CC=C(C=C1)C[C@@H](CO)N
TRPol: C1=CC=C2C(=C1)C(=CN2)C[C@@H](CO)N
AzAla: C1=CC=C2C=CC(=C2C=C1)C[C@@H](C(=O)O)N
DAP(Ac): C(C)(=O)NC(C(=O)O)CN
HLeu: CC(C)CC(C(=O)O)N
HSer: C(CO)C(C(=O)O)N
6-F-TRP: C1=CC2=C(C=C1F)NC=C2C[C@@H](C(=O)O)N
D-DAB: C(CN)[C@H](C(=O)O)N
2-Abz: C1=CC=C(C(=C1)C(=O)O)N
5-OH-TRP: C1=CC2=C(C=C1O)C(=CN2)CC(C(=O)O)N
HTrp: C1=CC=C2C(=C1)C(=CN2)CC(C(=O)O)N
Me-Phe: CN[C@@H](CC1=CC=CC=C1)C(=O)O
TERT BU PHE: CC(C)(C)C1=CC=C(C=C1)C[C@@H](C(=O)O)N
8-Aoc: C(CCCC(=O)O)CCCN
2-Aoc: CCCCCCC(C(=O)O)N
D-2-Aoc: N[C@@H](C(=O)O)CCCCCC
Me-Trp: CNC(CC1=CNC2=CC=CC=C21)C(=O)O
Pra: C#CCNCC(=O)O
HPhe: C1=CC=C(C=C1)CC[C@@H](C(=O)O)N
TERT BU SER: CC(C)(C)OC[C@@H](C(=O)O)N
D-4-Cl-PHE: C1=CC(=CC=C1C[C@H](C(=O)O)N)Cl
PHE(4-OMe): COC1=CC=C(C=C1)CC(C(=O)O)N
Me-Val: CC(C)C(C(=O)O)NC
Me-Asn: CN[C@@H](CC(N)=O)C(=O)O
D-Me-Leu: CC(C)C[C@H](C(=O)O)NC
Se-Cys: C([C@@H](C(=O)O)N)[Se]
D-4-HYP: C1[C@@H](CN[C@H]1C(=O)O)O
D-Me-Phe: CN[C@H](CC1=CC=CC=C1)C(=O)O
D-DAP: C([C@H](C(=O)O)N)N
LEU-Boro: B([C@H](CC(C)C)N)(O)O
Thi-ALA: C1=CSC(=C1)C[C@@H](C(=O)O)N
4-Cl-PHE: C1=CC(=CC=C1C[C@@H](C(=O)O)N)Cl
bLeu: CC(C)C(CC(=O)O)N
Me-Ala: C[C@@H](C(=O)O)NC
ACT-D-Orn: C(C)(=O)N[C@H](CCCN)C(=O)O
iPr-D-Orn: C(C)(C)N[C@H](CCCN)C(=O)O
4-NH2-Pro: C1[C@H](NCC1N)C(=O)O
Me-TYR: CNC(CC1=CC=C(C=C1)O)C(=O)O
Agb: C1C2CC3CC1CC(C2)(C3)NC(=O)NCC4=CC=C(C=C4)N=C(N)N
Et-Cys: CCSCC(C(=O)O)N
Anth-ALA: C1=CC=CC2=CC3=CC=CC=C3C(=C12)N[C@@H](C)C(=O)O
Me-CYS: CN[C@@H](CS)C(=O)O
Me-LYS: CNCCCC[C@@H](C(=O)O)N
Di-P-Tyr: C1=CC(=CC=C1C[C@@H](C(=O)O)N(P(=O)(O)O)P(=O)(O)O)O
NBu-Gly: CCCCNCC(=O)O
N1-NMe-Gly: C1=CC=C2C(=C1)C=CC=C2CNCC(=O)O
N4-MeBn-Gly: CC1=CC=C(C=C1)CNCC(=O)O
2-Adc: CCCCCCCCC(C(=O)O)N
HEtGly: OCCNCC(=O)O
FLG: C1C2=CC=CC=C2C3=C1C(=CC=C3)NCC(=O)O
NIle: CC(CC)NCC(=O)O
Nae: CCCCCCCCCCCC(=O)NCCO
NLeu: C(C(C)C)NCC(=O)O
Nai: C1C=CC2=CC(=CC=C12)NCC(=O)O
Nspe: C1=CC(=CC=C1[N+](=O)[O-])OP(=O)(CCCCC(=O)O)O
Act-Cys: C(C)(=O)N[C@@H](CS)C(=O)O
HCys: C(C(C(=O)O)N)S
D-Pra: C#CC[C@H](C(=O)O)N
(CH3)3-Gly: C[N+](C)(C)CC(=O)[O-]
N2H3-Gly: C(C(=O)NN)N
3F-PHE: C1=CC=C(C=C1)C([C@@](C(=O)O)(N)F)(F)F
NVA: CCCC(C(=O)O)N
Dpg: CCCC(CCC)(C(=O)O)N
5-NH2-C5: C(CCN)CC(=O)O
SER-O-BZY: C1=CC=C(C=C1)COCC(C(=O)O)N
D-AGL: N[C@H](CC=C)C(=O)O
2-Ahx: C/C=C/C[C@@H](C(=O)O)N
3-F-TYR: C1=CC(=C(C=C1CC(C(=O)O)N)F)O
2-F-TYR: FC1=C(C[C@H](N)C(=O)O)C=CC(=C1)O
Ntridec: C(CCCCCCCCCCCC)NCC(=O)O
Iaa: C1=CC=C2C(=C1)C(=CN2)CC(=O)O
Iac: C1=CC=C2C(=C1)C(=CN2)CC(=O)O
PHE(NH2): C1=CC(=CC=C1C[C@@H](C(=O)O)N)N
PHE(NH2-Bzl): NC1=C(C(=O)C2=CC=C(C[C@H](N)C(=O)O)C=C2)C=CC=C1
HPro: C1CC(NC1)C(=O)O
CM-Cys: C([C@@H](C(=O)O)N)SCC(=O)O
3-Abz: C1=CC(=CC(=C1)N)C(=O)O
D-4-F-PHE: FC1=CC=C(C[C@@H](N)C(=O)O)C=C1
CYS-C6: C(CCCCC)SC[C@H](N)C(=O)O
D-TYR-Bzl: C1=CC=C(C=C1)COC2=CC=C(C=C2)C[C@H](C(=O)O)N
CYS-Bzl: C1=CC=C(C=C1)CSC[C@@H](C(=O)O)N
Me-PRO: CN1[C@@H](CCC1)C(=O)O
Me-Thr: CN[C@@H]([C@H](O)C)C(=O)O
beta-Thi-D-ALA: C1=CSC(=C1)C[C@H](C(=O)O)N
THZ-Ala: C1=C(N=CS1)C[C@@H](C(=O)O)N
MIM: CC1=NC=CN1CCCCC2=CC=C(C=C2)CC(=O)N[C@@H](CO)C(=O)N[C@@H](CCCCN)C(=O)NCCC3CCCCC3
2-F-PHE: FC1=C(C[C@H](N)C(=O)O)C=CC=C1
N,N-Me-VAL: CC(C)C(C(=O)O)N(C)C
F-Pro: C1[C@H](NCC1F)C(=O)O
2F-Pro: C1[C@H](NCC1(F)F)C(=O)O
Chx-PRO: C1(CCCCC1)C1C[C@H](NC1)C(=O)O
Ph-Pro: C1(=CC=CC=C1)C1C[C@H](NC1)C(=O)O
Bn-Pro: C(C1=CC=CC=C1)C1C[C@H](NC1)C(=O)O
PhO-Pro: C1[C@H](NCC1OC2=CC=CC=C2)C(=O)O
N,N-Me-LEU: CN([C@@H](CC(C)C)C(=O)O)C
DIT: C1=C(C=C(C(=C1I)O)I)C[C@@H](C(=O)O)N
7-DMA-TRP: CC(=CCC1=C2C(=CC=C1)C(=CN2)C[C@@H](C(=O)O)N)C
Cbz-ORN: C1=CC=C(C=C1)COC(=O)N[C@@H](CCCN)C(=O)O
Cbz-LYS: C1=CC=C(C=C1)COC(=O)N[C@@H](CCCCN)C(=O)O
VALol: CC(C)C(C(=O)O)(N)O
N,O-Me-TYR: CN[C@@H](CC1=CC=C(C=C1)OC)C(=O)O
5-NH2-2-OH-C5: C(CC(C(=O)O)O)CN
O-Me-Tyr: COC1=CC=C(C=C1)CC(C(=O)O)N
3-I-Tyr: C1=CC(=C(C=C1C[C@@H](C(=O)O)N)I)O
3,5-I-Tyr: C1=C(C=C(C(=C1I)O)I)C[C@@H](C(=O)O)N
HTyr: C1=CC(=CC=C1CC[C@@H](C(=O)O)N)O
D-PHE(4-OMe): COC1=CC=C(C=C1)C[C@H](C(=O)O)N
IST: CC[C@H](C)[C@@H]([C@@H](CC(=O)O)O)N
N,O-Me-Ser: CN[C@@H](COC)C(=O)O
D-Me-Ala: C[C@H](C(=O)O)NC
2,5,7 TERT BU TRP : CC(C)(C)C1=CC2=C(C(=C1)C(C)(C)C)NC(=C2C[C@@H](C(=O)O)N)C(C)(C)C
OMe-Trp: CON[C@@H](CC1=CNC2=CC=CC=C12)C(=O)O
Asn(D-GlcNAc) : CC(=O)N[C@@H]1[C@H]([C@@H]([C@H](O[C@H]1NC(=O)C[C@@H](C(=O)O)N)CO)O)O
4-OH-Phg: C1=CC(=CC=C1[C@@H](C(=O)O)N)O
4-Abz: C1=CC(=CC=C1C(=O)O)N
THR-Cl: C(C(C(C(=O)O)N)O)Cl
5-Me-OPro: C[C@H]1C(=O)C[C@H](N1)C(=O)O
3-OH-5-Me-Pro: C[C@H]1C[C@@H]([C@H](N1)C(=O)O)O
THR-OH: C([C@H]([C@@H](C(=O)O)N)O)O
End: C1[C@H](NC(=N1)N)C[C@@H](C(=O)O)N
D-End: N[C@H](C(=O)O)CCCCCNC(=N)N
D-Phg: C1=CC=C(C=C1)[C@H](C(=O)O)N
bLYS: C(CC(CC(=O)O)N)CN
N,S-Me-CYS: CN[C@@H](CSC)C(=O)O
Allo-Thr: CC(C(C(=O)[O-])[NH3+])O
3-Hyp: C1CN[C@@H](C1O)C(=O)O
DHTrp: C1=CC=C2C(=C1)C(=CN2)/C=C(/C(=O)O)\N
3-OH-Leu: CC(C)C([C@@H](C(=O)O)N)O
D-Me-Ser: CN[C@H](CO)C(=O)O
Me-4-OH-PHG: CNC(C(=O)O)C1=CC=C(C=C1)O
3-NH2-C16: CCCCCCCCCCCCCC(CC(=O)O)N
Pen: CC(C)([C@H](C(=O)O)N)S
D-Pen: CC(C)([C@H](C(=O)O)N)S
D-Me-TYR: CN[C@@H](CC1=CC=C(C=C1)O)C(=O)O
P-Thr: C[C@H]([C@@H](C(=O)O)N)OP(=O)(O)O
Suc-Lys: C(CCN)C[C@@H](C(=O)O)NC(=O)CCC(=O)O
3-ME-GLU: CC(CC(=O)O)C(C(=O)O)N
KYN: C1=CC=C(C(=C1)C(=O)C[C@@H](C(=O)O)N)N
I3CA: C1=CC=C2C(=C1)C(=CN2)C=O
D-Pip: C1CCN[C@H](C1)C(=O)O
3-Me-ASP: C[C@@H]([C@H](C(=O)O)N)C(=O)O
Cl-Th2CA: C1[C@@H](O1)CNC(=O)C2=CC=C(S2)Cl
D-HSer: N[C@H](CCO)C(=O)O
3-OH-Asp: C(C(C(=O)O)O)(C(=O)O)N
11-NH2-C11: C(CCCCCN)CCCCC(=O)O
D-IGln: N[C@@H](C(=O)O)CC(N)=O
N5-Ac-N5-OH-ORN: CC(=O)N(CCC[C@@H](C(=O)O)N)O
Me-3-OH-Val: CN[C@@H](C(C)(C)O)C(=O)O
D-3-O-TRP: N1C=C(C2=CC=CC=C12)CC(C(=O)O)=O
D-3-OH-Asp: [C@@H](C(C(=O)O)O)(C(=O)O)N
Dab(betaOH): NC(C(=O)O)C(CN)O
3-OH-Ile: O[C@]([C@H](N)C(=O)O)(C)CC
N,N-Me-Phe: CN([C@H](C(=O)O)CC1=CC=CC=C1)C
N,N-Me-ALA: CN([C@@H](C)C(=O)O)C
4-Me-Pro: CC1C[C@H](NC1)C(=O)O
OH-Ile: CC[C@H](C)[C@@H](C(=O)O)NO
OH-VAL: ON[C@@H](C(C)C)C(=O)O
D-3-OH-VAL: CC(C)([C@H](C(=O)O)N)O
beta-HPhe: N[C@@H](CC1=CC=CC=C1)CC(=O)O
S-Me-CYS: CSC[C@@H](C(=O)O)N
LYS-C12: C(CCCCCCCCCCC)(=O)NCCCC[C@H](N)C(=O)O
LYS-C14: C(CCCCCCCCCCCCC)(=O)NCCCC[C@H](N)C(=O)O
LYS-C16: C(CCCCCCCCCCCCCCC)(=O)NCCCC[C@H](N)C(=O)O
LYS-ACT: C(C)(=O)NCCCC[C@H](N)C(=O)O
D-AIB: NC(C(=O)O)(C)C
Ser-GLC: C(C)(=O)N[C@H]1[C@@H](O[C@@H]([C@H]([C@@H]1O)O)CO)OC[C@H](N)C(=O)O
Thr-GLC : C(C)(=O)N[C@H]1[C@@H](O[C@@H]([C@H]([C@@H]1O)O)CO)O[C@@H]([C@H](N)C(=O)O)C
D-ABU: N[C@@H](C(=O)O)CC
NLYS: NCCCCNCC(=O)O
AGP: N(C(=N)N)CCC(=O)O
Me-ARG: N[C@H](C(=O)O)CCCN(C(=N)N)C
Asn-GLC : C(C)(=O)N[C@H]1[C@@H](O[C@@H]([C@H]([C@@H]1O)O)CO)NC(C[C@H](N)C(=O)O)=O
Acm-Cys: N[C@H](C(=O)O)CSCNC(C)=O
TERT BU ALA: NC(C(=O)O)C(C)(C)C
D-FLG: N[C@@H](C(=O)O)C1C2=CC=CC=C2C=2C=CC=CC12
4-NH2-D-PRO: N[C@@H]1[C@@H](NCC1)C(=O)O
CYS-S-CH3: CSSC[C@H](N)C(=O)O
S-ALA-4-pen: N[C@@](C(=O)O)(C)CCCC=C
R-ALA-7-oct: N[C@](C(=O)O)(C)CCCCCCC=C
CF3-Bpg: NC(C(=O)O)C12CC(C1)(C2)C(F)(F)F
D-CF3-Bpg: N[C@@H](C(=O)O)C12CC(C1)(C2)C(F)(F)F
bASP: NC(CC(=O)O)C(=O)O
AEt-VAL: NCCN[C@@H](C(C)C)C(=O)O
NNar: N(C(=N)N)CCCNCC(=O)O
CYS-PEG750: NC(C(=O)O)CSCCOCCOC
CYS-PEG2000: NC(C(=O)O)CSCCOCCOCCOC
Tic: C1NC(CC2=CC=CC=C12)C(=O)O
Oic: N1C(CC2CCCCC12)C(=O)O
D-LAP: C(CCCCCCCCCCC)N[C@H](CC(N)=O)C(=O)O
LYS-C8: C(CCCCCCC)(=O)NCCCC[C@H](N)C(=O)O
LYS-C4: C(CCC)(=O)NCCCC[C@H](N)C(=O)O
LYS-C12-OH: OCCCCCCCCCCCC(=O)NCCCC[C@H](N)C(=O)O
N,N-Me-Lys: CN(CCCC[C@H](N)C(=O)O)C
BIP: NC(C(=O)O)CC1=CC=C(C=C1)C1=CC=CC=C1
DAB(Ac): C(C)(=O)NC(C(=O)O)CCN
Ph-TRP: N[C@H](C(=O)O)CC1=C(NC2=CC=CC=C12)C1=CC=CC=C1
LYS-C5: C(CCCC)(=O)NCCCC[C@H](N)C(=O)O
LYS-Me-C5: CC(CCC(=O)NCCCC[C@H](N)C(=O)O)C
LYS-C6: C(CCCCC)(=O)NCCCC[C@H](N)C(=O)O
LYS-Me-C6: CC(C(=O)NCCCC[C@H](N)C(=O)O)CCCC
LYS-Me-C8: CC(CCC(=O)NCCCC[C@H](N)C(=O)O)CCCC
LYS-C18: C(CCCCCCCC=CCCCCCCCC)(=O)N[C@@H](CCCCN)C(=O)O
D-LYS-C4: C(CCC)(=O)NCCCC[C@@H](N)C(=O)O
D-LYS-C6: C(CCCCC)(=O)NCCCC[C@H](C(=O)O)N
D-LYS-Me-C5: CC(CCC(=O)NCCCC[C@@H](N)C(=O)O)C
D-LYS-C8: C(CCCCCCC)(=O)NCCCC[C@@H](N)C(=O)O
2-OH-Me-SER: OCN[C@H](C(=O)O)CO
10-Adc: NCCCCCCCCCC(=O)O
DIP: NC(C(=O)O)C(C1=CC=CC=C1)C1=CC=CC=C1
His[(CH2)3C6H5]2: C1(=CC=CC=C1)CCCN[C@H](C(=O)O)CC=1N=CN(C1)CCCC1=CC=CC=C1
Lys(N3): N[C@H](C(=O)O)CCCCN=[N+]=[N-]
LYS-C10: C(CCCCCCCCC)(=O)NCCCC[C@H](N)C(=O)O
12-Guan-C12: N(C(=N)N)CCCCCCCCCCCC(=O)O
MET-Boro: NC(CB(O)O)CCSC
TERT BU CYS: N[C@H](C(=O)O)CSC(C)(C)C
D-Thr-GLC: C(C)(=O)N[C@H]1[C@@H](O[C@@H]([C@H]([C@@H]1O)O)CO)O[C@H]([C@@H](N)C(=O)O)C
N,N-Me-D-Orn: N[C@@H](C(=O)O)CCCN(C)C
N,N,N-Me-D-Orn: N[C@@H](C(=O)[O-])CCC[N+](C)(C)C
MOR-D-Orn: N[C@@H](C(=O)O)CCCN1CCOCC1
Hex-Hyp: C(CCCCC)O[C@@H]1C[C@H](NC1)C(=O)O
Dab(Cys): N[C@@H](CS)C(=O)N[C@H](C(=O)O)CCN
CYS-S-HexNAc: N[C@H](C(=O)O)CS[C@H]1[C@@H]([C@@H](O)[C@H](O)[C@H](O1)CO)NC(C)=O
bHTrp: NC(C(=O)O)CCC1=CNC2=CC=CC=C12
4-3Fm-PHE: NC(C(=O)O)CC1=CC=C(C=C1)C(F)(F)F
D-BIP: N[C@@H](C(=O)O)CC1=CC=C(C=C1)C1=CC=CC=C1
D-Me-Trp: N[C@@H](C(=O)O)CC1=CN(C2=CC=CC=C12)C
Me-PENT-GLY: N[C@](C(=O)O)(CCCC=C)C
LYS-C7: C(CCCCCC)(=O)NCCCC[C@H](N)C(=O)O
Lys-C9: C(CCCCCCCC)(=O)NCCCC[C@H](N)C(=O)O
Lys-C11: C(CCCCCCCCCC)(=O)NCCCC[C@H](N)C(=O)O
D-LYS-C10: C(CCCCCCCCC)(=O)NCCCC[C@@H](N)C(=O)O
D-LYS-C12: C(CCCCCCCCCCC)(=O)NCCCC[C@@H](N)C(=O)O
D-LYS-C14: C(CCCCCCCCCCCCC)(=O)NCCCC[C@@H](N)C(=O)O
D-LYS-C16: C(CCCCCCCCCCCCCCC)(=O)NCCCC[C@@H](N)C(=O)O
CYS-C12: C(CCCCCCCCCCC)(=O)SC[C@H](N)C(=O)O
4-Br-PHE: NC(C(=O)O)CC1=CC=C(C=C1)Br
4-I-PHE: NC(C(=O)O)CC1=CC=C(C=C1)I
LYS-NH2-C5: NCCCCC(=O)NCCCC[C@H](N)C(=O)O
LYS-Ahx: NCCCCCC(=O)NCCCC[C@H](N)C(=O)O
LYS-BrC6: BrCCCCCC(=O)NCCCC[C@H](N)C(=O)O
LYS-NH2-C11: NCCCCCCCCCCC(=O)NCCCC[C@H](N)C(=O)O
NTrp: N1C=C(C2=CC=CC=C12)CCNCC(=O)O
Ndpe: C1(=CC=CC=C1)C(CNCC(=O)O)C1=CC=CC=C1
Ala-d3: N[C@H](C(=O)O)C([2H])([2H])[2H]
IAA-Cys: N[C@H](C(=O)O)CSCCC(N)=O
5F-PHE: NC(C(=O)O)CC1=C(C(=C(C(=C1F)F)F)F)F
DAP-C10: C(CCCCCCCCC)(=O)NCC(C(=O)O)N
DAP-C12: C(CCCCCCCCCCC)(=O)NCC(C(=O)O)N
DAP-C14: C(CCCCCCCCCCCCC)(=O)NCC(C(=O)O)N
DAP-C16: C(CCCCCCCCCCCCCCC)(=O)NCC(C(=O)O)N
DAB-C10: C(CCCCCCCCC)(=O)NCCC(C(=O)O)N
DAB-C12: C(CCCCCCCCCCC)(=O)NCCC(C(=O)O)N
DAB-C14: C(CCCCCCCCCCCCC)(=O)NCCC(C(=O)O)N
DAB-C16: C(CCCCCCCCCCCCCCC)(=O)NCCC(C(=O)O)N
His[(CH2)7C6H5]2: C1(=CC=CC=C1)CCCCCCCN[C@H](C(=O)O)CC=1N=CN(C1)CCCCCCCC1=CC=CC=C1
His[(CH2)7C6H5]: C1(=CC=CC=C1)CCCCCCCN[C@@H](CC1=CNC=N1)C(=O)O
HCha: NC(C(=O)O)CCCC1CCCCC1
Leu(O)S: NC(C(O)=S)CC(C)C
BisHomo-Pra: NC(C(=O)O)CCC#C
N,4-Me-Glu	: CN[C@@H](CC(C(=O)O)C)C(=O)O
Pre-TRP: N[C@H](C(=O)O)CC1=CN(C2=CC=CC=C12)CC=C(C)C
N-TYR: NC(C(=O)O)CC1=CC=C(C=C1)[N+](=O)[O-]
R-ALA-4-pen: N[C@@H](C(=O)O)CCCCC=C
Glu-OAll: N[C@H](C(=O)OCC=C)CCC(=O)[O-]
LYS-PEG4: COCCOCCNCCCC[C@H](N)C(=O)O
LYS-C24: C(CCCCCCCCCCCCCCCCCCCCCCC)(=O)NCCCC[C@H](N)C(=O)O
LYS-C22: C(CCCCCCCCCCCCCCCCCCCCC)(=O)NCCCC[C@H](N)C(=O)O
LYS-C20: C(CCCCCCCCCCCCCCCCCCC)(=O)NCCCC[C@H](N)C(=O)O
LYS-PEG8: COCCOCCOCCOCCNCCCC[C@H](N)C(=O)O
S-Fmoc-ALA-4-pen: C1=CC=CC=2C3=CC=CC=C3C(C12)COC(=O)N[C@H](C(=O)O)CCCCC=C
Me-Ile: CN[C@@H]([C@@H](C)CC)C(=O)O
BTA: NC(C(=O)O)CC1=CSC2=C1C=CC=C2
DAB-C4: C(CCC)(=O)NCCC(C(=O)O)N
DAB-C6: C(CCCCC)(=O)NCCC(C(=O)O)N
DAB-C8: C(CCCCCCC)(=O)NCCC(C(=O)O)N
D-Gly-C8: N[C@@H](C(=O)O)CCCCCCCC
Gly-C8: NC(C(=O)O)CCCCCCCC
CYS-PEG4-Chol: NC(C(=O)O)CSCCOCCO[C@@H]1CC2=CC[C@H]3[C@@H]4CC[C@H]([C@@H](CCCC(C)C)C)[C@]4(CC[C@@H]3[C@]2(CC1)C)C
Eps-LYS: NCCCCCC(=O)O
S-TERT BU CYS: N[C@H](C(=O)O)CSSC(C)(C)C
beta-BzThi-Ala: NC(C(=O)O)CC1=CSC2=C1C=CC=C2
CYS-C16: C(CCCCCCCCCCCCCCC)(=O)SC[C@H](N)C(=O)O
2F(3,5)-PHE: NC(C(=O)O)CC1=CC(=CC(=C1)F)F
PHE-NH2-C5: NC(C(=O)O)CC1=CC=C(C=C1)C(CCCCN)=O
ORN-NH2-C5: NCCCCC(=O)NCCC[C@H](N)C(=O)O
ORN-NH2-C7: NCCCCCCC(=O)NCCC[C@H](N)C(=O)O
2F(2,6)-PHE: NC(C(=O)O)CC1=C(C=CC=C1F)F
6F-LEU: NC(C(=O)O)CC(C(F)(F)F)(F)F
CF3-Pro: FC([C@@]1(NCCC1)C(=O)O)(F)F
CH3-S-Pro: CS[C@@H]1C[C@H](NC1)C(=O)O
CYS-PEG5000: NC(C(=O)O)CSCCOCCOC
mPEG-2000-Mal-CYS: N[C@@H](C(=O)O)CN1C(C(=CC1=O)CCOCCOC)=O
D-Lys(N3): N[C@@H](C(=O)O)CCCCN=[N+]=[N-]
2-DMA-TRP: N[C@H](C(=O)O)CC1=C(NC2=CC=CC=C12)C(=C(C)C)C
Cbz-D-ORN: C(C1=CC=CC=C1)OC(=O)NCCC[C@@H](N)C(=O)O
LYS-PA: N[C@H](C(=O)O)CCCCNC(C=CC=CC1=CC2=C(C=C1)OCO2)=O
ORN-PA: N[C@H](C(=O)O)CCCNC(C=CC=CC1=CC2=C(C=C1)OCO2)=O
LYS-EPA: N[C@H](C(=O)O)CCCCNC(C=CC=CC1=CC2C(C=C1)(OCO2)CC)=O
ORN-EPA: N[C@H](C(=O)O)CCCNC(C=CC=CC1=CC2C(C=C1)(OCO2)CC)=O
CYS-PEG12-Chol: NC(C(=O)O)CSCCOCCOCCOCCOCCOCCOCCOCCOCCOCCOCCOCCO[C@@H]1CC2=CC[C@H]3[C@@H]4CC[C@H]([C@@H](CCCC(C)C)C)[C@]4(CC[C@@H]3[C@]2(CC1)C)C
D-3-OH-ASN: N[C@@H](C(=O)O)[C@@H](O)N
CYS-Chol: N[C@H](C(=O)O)CS[C@@H]1CC2=CC[C@H]3[C@@H]4CC[C@H]([C@@H](CCCC(C)C)C)[C@]4(CC[C@@H]3[C@]2(CC1)C)C
CYS-PEG24-Chol: NC(C(=O)O)CSCCOCCOCCOCCOCCOCCOCCOCCOCCOCCOCCOCCO CCOCCOCCOCCOCCOCCOCCOCCOCCOCCOCCO[C@@H]1CC2=CC[C@H]3[C@@H]4CC[C@H]([C@@H](CCCC(C)C)C)[C@]4(CC[C@@H]3[C@]2(CC1)C)C
5-Cl-TRP: NC(C(=O)O)CC1=CNC2=CC=C(C=C12)Cl
D-bALA: NCCC(=O)O
AMiPPA: NC(C(=O)O)CC(CC(C(=O)O)CC(C)O)C
MPro: S[C@@H]1C[C@H](NC1)C(=O)O
Dab(Thr): N[C@H](C(=O)N[C@H](C(=O)O)CCN)[C@@H](C)O
D-Dab(Thr): N[C@H](C(=O)N[C@@H](C(=O)O)CCN)[C@@H](C)O
3-Me-Trp: NC(C(=O)O)CC1(CNC2=CC=CC=C12)C
OMe-Glu: COC(CC[C@@H](C(=O)O)N)=O
3-NH2-2-OH-4,15-Me-(E)14-C19: NC(C(C(=O)O)O)CCCCCCCCCCC=C(CCCC)C
2-NH2-2,6,8-Me-C10: NC(C(=O)O)(CCC(C(CCCC)C)C)C
D-OMe-Glu: COC(CC[C@H](C(=O)O)N)=O
N1,N2-Me-ORN: CN([C@@H](CCCN)C(=O)O)C
SER-O-GlcNAc: C(C)(=O)N[C@H]1[C@@H](O[C@@H]([C@H]([C@@H]1O)O)CO)OC[C@H](N)C(=O)O
4-F-ABU: NC(C(=O)O)CCF
4-2F-ABU: NC(C(=O)O)CC(F)F
4-3F-ABU: NC(C(=O)O)CC(F)(F)F
4-2F-5-3F-NVA: NC(C(=O)O)CC(C(F)(F)F)(F)F
D-SER-O-OMe-CIN: COC1=CC=C(C=CC(=O)OC[C@@H](N)C(=O)O)C=C1
D-SER-O-Me-CIN: CC1=CC=C(C=CC(=O)OC[C@@H](N)C(=O)O)C=C1
SER-O-Me-CIN: CC1=CC=C(C=CC(=O)OC[C@H](N)C(=O)O)C=C1
D-Piz: N1[C@H](CNCC1)C(=O)O
DHLys: N[C@H](C(=O)O)CCC=C
13-N,N-Me-Guan-(E)2,4,8-C13: CN(C(NC(CC=CCC=CCC=CCCC(=O)O)CC)=N)C
LYS-Bu: C(CC#C)(=O)NCCCC[C@H](N)C(=O)O
DAP-Bu: C(CC#C)(=O)NCC(C(=O)O)N
\end{lstlisting}
\section{Genome Information}\label{secC}

\textbf{All the genome from ATCC:}
\begin{lstlisting}
Aggregatibacter actinomycetemcomitans ATCC 29523
Burkholderia multivorans ATCC 17616
Gardnerella vaginalis ATCC 49145
Enterobacter cloacae ATCC BAA 2468
Salmonella enterica subsp enterica serovar Typhimurium ATCC 14028
Yersinia enterocolitica subsp enterocolitica ATCC 23715
Pseudomonas fluorescens ATCC 13525
Streptococcus uberis ATCC 19436
Escherichia coli ATCC #005
Enterobacter cloacae subsp cloacae ATCC 23355
Aeromonas veronii ATCC 35624
Bordetella pertussis ATCC 9340
Clostridioides difficile ATCC 43598
Escherichia coli ATCC 23506
Stenotrophomonas maltophilia ATCC 51331
Shigella sonnei ATCC 11060
Enterococcus faecalis ATCC 33186
Pseudomonas aeruginosa ATCC BAA 1744
Alloprevotella tannerae ATCC 51259
Micrococcus luteus ATCC 4698
Aspergillus awamori ATCC 22342
Aeromonas hydrophila ATCC 49140
Bacillus pacificus ATCC 10987
Klebsiella pneumoniae subsp pneumoniae ATCC 12657
Priestia megaterium ATCC 19213
Staphylococcus aureus ATCC 25904
Enterococcus faecalis ATCC 19433
Escherichia coli ATCC 35218
Fusarium oxysporum ATCC MYA 1198
Staphylococcus aureus subsp aureus ATCC BAA 39
Clostridium perfringens ATCC 3624
Streptococcus sanguinis ATCC 10556
Streptococcus pyogenes ATCC 12344
Escherichia coli ATCC BAA 1025
Haemophilus influenzae ATCC 51907
Cutibacterium acnes ATCC 11828
Neisseria gonorrhoeae ATCC 49226
Klebsiella aerogenes ATCC 51697
Cronobacter sakazakii ATCC BAA 894
Morganella morganii subsp morganii ATCC 29853
Staphylococcus aureus ATCC BAA 1683
Candida tropicalis ATCC 750
Lactobacillus plantarum subsp plantarum ATCC 14917
Mycoplasma hyorhinis ATCC 17981
Clostridium sporogenes ATCC 19404
Legionella pneumophila subsp pneumophila ATCC 33215
Streptococcus pneumoniae ATCC 6303
Listonella anguillarum ATCC 19264
Candida glabrata ATCC 2001
Acinetobacter baumannii ATCC BAA 1798
Aerococcus viridans ATCC 700406
Saccharomyces cerevisiae ATCC 13007
Mycobacterium tuberculosis variant bovis BCG ATCC 35734
Staphylococcus aureus ATCC BAA 1720
Klebsiella pneumoniae subsp pneumoniae ATCC BAA 2524
Bacillus atrophaeus ATCC 9372
Saccharomyces cerevisiae ATCC 204508
Aeromonas sobria ATCC 43979
Escherichia coli ATCC 11229
Staphylococcus aureus ATCC 13565
Pseudomonas aeruginosa ATCC 25668
Escherichia coli ATCC 51659
Trichophyton interdigitale ATCC 9533
Streptococcus pneumoniae ATCC 33400
Enterococcus faecium ATCC 27270
Aeromonas hydrophila ATCC 7966
Helicobacter pylori ATCC 43504
Staphylococcus carnosus subsp carnosus ATCC 51365
Bacillus cereus ATCC 10876
Streptococcus pneumoniae ATCC BAA 255
Moraxella catarrhalis ATCC 8176
Priestia megaterium ATCC 10778
Staphylococcus lugdunensis ATCC 43809
Aspergillus brasiliensis ATCC 9642
Klebsiella aerogenes ATCC 49701
Pseudomonas aeruginosa ATCC 19660
Bacillus spizizenii ATCC 6633
Enterococcus faecalis ATCC 47077
Cutibacterium acnes ATCC 51277
Pseudomonas aeruginosa ATCC 9721
Candida albicans ATCC 24433
Pseudomonas aeruginosa ATCC 25619
Aspergillus terreus ATCC 1012
Fusobacterium nucleatum subsp nucleatum ATCC 25586
Candida kefyr ATCC 66028
Acinetobacter baumannii ATCC BAA 1794
Staphylococcus epidermidis ATCC 700565
Fusobacterium nucleatum subsp nucleatum ATCC 23726
Vibrio sp ATCC 19108
Bacteroides fragilis ATCC 25285
Klebsiella pneumoniae subsp pneumoniae ATCC 27799
Priestia megaterium ATCC 14581
Candida albicans ATCC MYA 2876
Micrococcus luteus ATCC 49732
Pseudomonas aeruginosa ATCC #002
Salmonella enterica subsp enterica serovar Choleraesuis ATCC 13312
Cutibacterium acnes ATCC 29399
Lactobacillus crispatus ATCC 33820
Bacillus amyloliquefaciens ATCC 23842
Torulaspora delbrueckii ATCC 10662
Haemophilus influenzae ATCC 19418
Mycobacterium tuberculosis subsp tuberculosis ATCC 27294
Saccharomyces cerevisiae ATCC 9763
Streptococcus mutans ATCC 700610
Malassezia sympodialis ATCC 42132
Pseudomonas aeruginosa ATCC 15442
Klebsiella pneumoniae subsp pneumoniae ATCC 13883
Micrococcus luteus ATCC 10240
Escherichia coli ATCC #001
Salmonella enterica subsp enterica serovar Enteritidis ATCC 13076
Salmonella enterica subsp enterica serovar Typhimurium ATCC 43971
Streptococcus sobrinus ATCC 33478
Proteus mirabilis ATCC 21100
Lactococcus lactis subsp lactis ATCC 19435
Lacticaseibacillus casei ATCC 393
Lactobacillus fermentum ATCC 9338
Candida albicans ATCC 76485
Romboutsia lituseburensis ATCC 25759
Pseudomonas aeruginosa ATCC 10145
Vibrio fluvialis ATCC 33809
Klebsiella michiganensis ATCC 43086
Bacteroides fragilis ATCC 29771
Aspergillus tubingensis ATCC 1004
Pseudomonas aeruginosa ATCC 47085
Pseudomonas aeruginosa ATCC BAA 2110
Staphylococcus warneri ATCC 49454
Acinetobacter lwoffii ATCC 15309
Xanthomonas citri pathovar mangiferaeindicae ATCC 11637
Enterobacter hormaechei ATCC 700323
Streptococcus sobrinus ATCC 27352
Staphylococcus aureus ATCC BAA 1718
Citrobacter freundii ATCC 8090
Escherichia coli ATCC 47076
Staphylococcus aureus ATCC 49525
Staphylococcus epidermidis ATCC 35984
Staphylococcus aureus ATCC 43300
Raoultella ornithinolytica ATCC 31898
Enterococcus faecium ATCC 35667
Bacillus spizizenii ATCC 19659
Aspergillus flavus ATCC 9643
Helicobacter pylori ATCC 43629
Mycobacterium kansasii ATCC 12478
Proteus hauseri ATCC 13315
Escherichia coli ATCC #004
Aspergillus fumigatus ATCC 96918
Escherichia coli ATCC 23716
Serratia sp ATCC 39006
Salmonella enterica subsp enterica serovar Typhimurium ATCC 700408
Serratia marcescens ATCC 17991
Burkholderia cepacia ATCC 25416
Listeria monocytogenes ATCC 7644
Shigella dysenteriae ATCC 13313
Campylobacter jejuni subsp jejuni ATCC 33291
Enterococcus faecium ATCC 51559
Nocardia farcinica ATCC 3308
Burkholderia cepacia ATCC 17765
Staphylococcus aureus ATCC 33591
Xanthomonas vesicatoria ATCC 35937
Bacteroides thetaiotaomicron ATCC 29741
Candida albicans ATCC 18804
Klebsiella pneumoniae subsp pneumoniae ATCC 10031
Fusobacterium varium ATCC 27725
Staphylococcus aureus ATCC BAA 1717
Listeria welshimeri ATCC 35897
Salmonella enterica subsp diarizonae ATCC 29934
Fusarium keratoplasticum ATCC 36031
Escherichia coli ATCC 23724
Bacillus cereus ATCC 14579
Listeria monocytogenes ATCC 19115
Clostridioides difficile ATCC 43255
Proteus mirabilis ATCC 29906
Pseudomonas aeruginosa ATCC 29260
Rhodotorula mucilaginosa ATCC 9449
Salmonella enterica subsp enterica serovar Typhimurium ATCC 700720
Listeria monocytogenes ATCC 19113
Enterococcus faecalis ATCC BAA 2365
Pectobacterium carotovorum ATCC 15713
Vibrio parahaemolyticus ATCC 17802
Bacillus licheniformis ATCC 14580
Saccharomyces kudriavzevii ATCC 2601
Parvimonas micra ATCC 33270
Streptococcus gallolyticus ATCC 9809
Shigella sonnei ATCC 9290
Streptococcus dysgalactiae subsp dysgalactiae ATCC 27957
Pseudomonas aeruginosa ATCC 27853
Legionella micdadei ATCC 33218
Serratia marcescens ATCC 8100
Listeria monocytogenes ATCC 19118
Acinetobacter junii ATCC 17908
Staphylococcus aureus ATCC 29737
Escherichia coli ATCC 31616
Lactococcus lactis subsp lactis ATCC 11454
Aspergillus fijiensis ATCC 20611
Aspergillus brasiliensis ATCC 16404
Klebsiella pneumoniae ATCC BAA 2146
Proteus vulgaris ATCC 6380
Escherichia coli ATCC 15597
Bifidobacterium longum subsp infantis ATCC 15697
Staphylococcus epidermidis ATCC 14990
Streptococcus pneumoniae ATCC 700677
Klebsiella pneumoniae subsp pneumoniae ATCC BAA 2470
Salmonella enterica subsp enterica serovar Typhimurium ATCC 29629
Ralstonia solanacearum ATCC 11696
Candida tropicalis ATCC 66029
Escherichia coli ATCC BAA 2471
Rhodococcus equi ATCC 6939
Staphylococcus epidermidis ATCC 51625
Streptococcus agalactiae ATCC 27956
Salmonella enterica subsp arizonae ATCC 13314
Helicobacter pylori ATCC 51653
Peptoniphilus asaccharolyticus ATCC 29743
Staphylococcus aureus ATCC BAA 44
Listeria innocua ATCC 33090
Bacillus subtilis ATCC 21332
Candida dubliniensis ATCC MYA 646
Shigella flexneri ATCC 29903
Pseudomonas aeruginosa ATCC 35032
Bifidobacterium longum ATCC BAA 999
Prevotella intermedia ATCC 49046
Enterobacter cloacae subsp cloacae ATCC 35030
Legionella bozemanae ATCC 33217
Staphylococcus aureus ATCC 700699
Mycobacterium marinum ATCC 927
Escherichia coli ATCC #003
Vibrio cholerae ATCC 25873
Mycobacterium vaccae ATCC 15483
Moraxella catarrhalis ATCC 25238
Streptococcus oralis ATCC 35037
Streptococcus gordonii ATCC 35105
Streptococcus pyogenes ATCC 14289
Enterococcus faecalis ATCC 29212
Escherichia coli ATCC 35150
Streptococcus dysgalactiae subsp equisimilis ATCC 12388
Klebsiella pneumoniae ATCC BAA 2814
Streptococcus salivarius subsp salivarius ATCC 7073
Streptococcus mutans ATCC 25175
Enterococcus durans ATCC 19432
Legionella pneumophila subsp pneumophila ATCC 33155
Escherichia coli ATCC 700926
Enterobacter cloacae subsp cloacae ATCC BAA 1143
Bacillus cereus ATCC 11778
Candida albicans ATCC 11006
Streptococcus salivarius subsp salivarius ATCC 25975
Streptococcus pyogenes ATCC 10389
Shigella sonnei ATCC 29930
Streptococcus pneumoniae ATCC 27336
Staphylococcus aureus subsp aureus ATCC 49775
Salmonella enterica subsp enterica ATCC 51741
Acinetobacter baumannii ATCC BAA 1800
Salmonella enterica subsp enterica serovar Typhi ATCC 700931
Escherichia coli ATCC 700928
Staphylococcus epidermidis ATCC 12228
Streptococcus agalactiae ATCC 13813
Candida albicans ATCC 2091
Streptococcus pyogenes ATCC 49399
Candida albicans ATCC 14053
Escherichia coli ATCC 43894
Staphylococcus aureus ATCC 29213
Edwardsiella ictaluri ATCC 33202
Listeria monocytogenes ATCC BAA 679
Listeria monocytogenes ATCC 19112
Candida krusei ATCC 14243
Veillonella parvula ATCC 10790
Salmonella enterica subsp enterica serovar Paratyphi A ATCC 9150
Staphylococcus aureus ATCC BAA 1708
Aspergillus flavus ATCC 204304
Meyerozyma guilliermondii ATCC 6260
Listeria monocytogenes ATCC 35152
Stenotrophomonas maltophilia ATCC 13636
Escherichia coli ATCC 15224
Cutibacterium acnes ATCC 11827
Aeromonas hydrophila ATCC 35654
Escherichia coli ATCC 9637
Staphylococcus aureus ATCC 6538P
Streptococcus pyogenes ATCC 19615
Cryptococcus neoformans var grubii ATCC 208821
Staphylococcus epidermidis ATCC 29887
Candida albicans ATCC 90028
Escherichia coli ATCC BAA 3170
Staphylococcus simulans ATCC 27848
Vibrio splendidus ATCC 33125
Staphylococcus aureus ATCC 12598
Blautia coccoides ATCC 29236
Streptococcus oralis ATCC 9811
Candida parapsilosis ATCC 22019
Escherichia coli ATCC 11775
Raoultella terrigena ATCC 33257
Pasteurella aerogenes ATCC 27883
Alternaria alternata ATCC 66981
Aspergillus niger ATCC 6275
Streptococcus pneumoniae ATCC 6302
Staphylococcus aureus ATCC 700698
Proteus vulgaris ATCC 49132
Streptococcus pneumoniae ATCC 49619
Clostridium nexile ATCC 27757
Escherichia coli ATCC 4157
Klebsiella pneumoniae ATCC BAA 1705
Klebsiella oxytoca ATCC 700324
Klebsiella pneumoniae subsp pneumoniae ATCC 33495
Candida tropicalis ATCC 13803
Legionella pneumophila subsp fraseri ATCC 33216
Staphylococcus aureus ATCC 33593
Proteus mirabilis ATCC 25933
Legionella dumoffii ATCC 33279
Lactobacillus rhamnosus ATCC 8530
Pseudomonas putida ATCC 12633
Trichophyton rubrum ATCC 28188
Acidipropionibacterium acidipropionici ATCC 4875
Pseudomonas aeruginosa ATCC 12121
Staphylococcus aureus ATCC BAA 1707
Aspergillus niger ATCC 9142
Chromobacterium subtsugae ATCC 31532
Acinetobacter baumannii ATCC BAA 1710
Burkholderia thailandensis ATCC 700388
Corynebacterium renale ATCC 19412
Propionibacterium freudenreichii subsp shermanii ATCC 9614
Fusobacterium nucleatum subsp vincentii ATCC 49256
Escherichia coli ATCC 700336
Salmonella enterica subsp enterica serovar Choleraesuis ATCC 10708
Enterococcus faecium ATCC 6569
Enterobacter cloacae ATCC BAA 3045
Staphylococcus aureus ATCC BAA 2312
Alicyclobacillus acidoterrestris ATCC 49025
Neisseria gonorrhoeae ATCC 43069
Escherichia coli ATCC 11303
Cutibacterium acnes ATCC 33179
Staphylococcus saprophyticus ATCC BAA 750
Lactobacillus gasseri ATCC 33323
Salmonella enterica subsp enterica ATCC 39183
Klebsiella oxytoca ATCC 49131
Acinetobacter baumannii ATCC 17978
Staphylococcus aureus ATCC BAA 41
Escherichia coli ATCC 8739
Candida parapsilosis ATCC 90018
Staphylococcus aureus ATCC 12600
Mycobacterium tuberculosis ATCC 25177
Bacillus cereus ATCC 13472
Cryptococcus neoformans var grubii ATCC 13690
Staphylococcus haemolyticus ATCC 29970
Staphylococcus aureus ATCC BAA 1696
Enterococcus faecium ATCC 19434
Listeria monocytogenes ATCC 19114
Acinetobacter baumannii ATCC BAA 1605
Staphylococcus aureus ATCC 35556
Escherichia coli ATCC 27325
Klebsiella pneumoniae subsp pneumoniae ATCC 43816
Bacillus thuringiensis ATCC 10792
Candida albicans ATCC 66027
Escherichia coli ATCC 43827
Alcaligenes faecalis subsp faecalis ATCC 8750
Escherichia coli ATCC 25922
Escherichia coli ATCC 10536
Salmonella enterica subsp enterica serovar Typhimurium ATCC 13311
Kocuria rhizophila ATCC 9341
Serratia marcescens ATCC 274
Shewanella haliotis ATCC 49138
Fusobacterium nucleatum subsp polymorphum ATCC 10953
Streptococcus mutans ATCC 35668
Vibrio vulnificus ATCC 27562
Listeria monocytogenes ATCC 15313
Bifidobacterium adolescentis ATCC 15703
Legionella longbeachae ATCC 33484
Lactobacillus fermentum ATCC 14931
Staphylococcus aureus ATCC 14458
Citrobacter koseri ATCC BAA 895
Klebsiella pneumoniae subsp pneumoniae ATCC 35657
Pseudomonas putida ATCC 49128
Klebsiella pneumoniae ATCC BAA 2342
Escherichia coli ATCC 25404
Listeria grayi ATCC 25401
Cutibacterium acnes ATCC 6919
Enterobacter ludwigii ATCC 49141
Staphylococcus aureus subsp aureus ATCC BAA 2762
Pseudomonas syringae pathovar tomato ATCC BAA 871
Escherichia coli ATCC 700728
Listeria ivanovii subsp ivanovii ATCC 19119
Bacillus subtilis ATCC 11774
Staphylococcus aureus ATCC 6538
Aspergillus fumigatus ATCC 1022
Haemophilus influenzae ATCC 49766
Pasteurella multocida subsp multocida ATCC 43137
Corynebacterium xerosis ATCC 373
Pseudomonas aeruginosa ATCC 15692
Legionella oakridgensis ATCC 33761
Clostridium perfringens ATCC 3626
Serratia marcescens subsp marcescens ATCC 13880
Mycobacterium abscessus ATCC 19977
Streptococcus equi subsp zooepidemicus ATCC 43079
Pseudomonas aeruginosa ATCC 7700
Helicobacter pylori ATCC 700392
Pseudomonas aeruginosa ATCC BAA 427
Corynebacterium pseudotuberculosis ATCC 19410
Phocaeicola vulgatus ATCC 8482
Staphylococcus aureus ATCC 25923
Aeromonas salmonicida subsp salmonicida ATCC 33658
Rhizobium radiobacter ATCC 15955
Paracoccidioides lutzii ATCC MYA 826
Streptococcus oralis ATCC 6249
Staphylococcus aureus ATCC 27660
Lactococcus cremoris ATCC 19257
Klebsiella pneumoniae ATCC BAA 1706
Salmonella enterica subsp enterica serovar Typhi ATCC 6539
Candida lusitaniae ATCC 34449
Acinetobacter baumannii ATCC 17961
Ruminococcus gnavus ATCC 29149
Streptococcus gordonii ATCC 10558
Klebsiella pneumoniae subsp pneumoniae ATCC 49472
Clostridium perfringens ATCC 10543
Lactobacillus jensenii ATCC 25258
Mycobacterium smegmatis ATCC 607
Helicobacter pylori ATCC 43526
Mycobacterium smegmatis ATCC 14468
Pseudomonas aeruginosa ATCC BAA 2108
Escherichia coli ATCC 12014
Bacillus paranthracis ATCC 13061
Aspergillus fumigatus ATCC 204305
Pseudomonas aeruginosa ATCC BAA 3197
Klebsiella pneumoniae subsp pneumoniae ATCC 27736
Aspergillus niger ATCC 1015
Mycobacterium fortuitum subsp fortuitum ATCC 6841
Shigella sonnei ATCC 25931
Exiguobacterium chiriqhucha ATCC 49676
Staphylococcus aureus subsp aureus ATCC 19636
Staphylococcus aureus ATCC BAA 1750
Lactobacillus vaginalis ATCC 49540
Candida albicans ATCC 96901
Pseudomonas aeruginosa ATCC BAA 2111
Staphylococcus hominis subsp hominis ATCC 27844
Xanthomonas phaseoli ATCC 9563
Klebsiella pneumoniae subsp pneumoniae ATCC 13882
Gardnerella vaginalis ATCC 14018
Enterobacter cloacae subsp cloacae ATCC 13047
Weizmannia coagulans ATCC 7050
Lactobacillus acidophilus ATCC 4356
Yersinia ruckeri ATCC 29473
Klebsiella pneumoniae subsp pneumoniae ATCC 23356
Escherichia coli ATCC BAA 2452
Actinomyces naeslundii ATCC 12104
Candida albicans ATCC 200955
Bacillus tropicus ATCC 4342
Candida tropicalis ATCC MYA 3404
Cryptococcus neoformans var grubii ATCC 90112
Ogataea polymorpha ATCC 34438
Listeria monocytogenes ATCC 13932
Corynebacterium jeikeium ATCC BAA 949
Yersinia enterocolitica subsp enterocolitica ATCC 9610
Mycobacterium avium subsp avium ATCC 15769
Propionibacterium freudenreichii subsp freudenreichii ATCC 6207
Escherichia coli ATCC BAA 2469
Haemophilus influenzae ATCC 10211
Pseudomonas aeruginosa ATCC 13388
Enterococcus casseliflavus ATCC 700327
Staphylococcus aureus ATCC 9144
Stenotrophomonas maltophilia ATCC 13637
Streptococcus iniae ATCC 29178
Staphylococcus saprophyticus ATCC 49907
Obesumbacterium proteus ATCC 25927
Proteus mirabilis ATCC 35659
Streptoverticillium verticillium subsp tsukushiense ATCC 21633
Haemophilus influenzae ATCC 49247
Lactiplantibacillus plantarum ATCC 8014
Pseudomonas aeruginosa ATCC BAA 2114
Campylobacter jejuni subsp jejuni ATCC 33292
Acinetobacter baumannii ATCC 19606
Listeria grayi ATCC 19120
Ruminococcus torques ATCC 27756
Serratia marcescens ATCC 21074
Enterococcus faecalis ATCC 14506
Enterococcus faecium ATCC 700221
Ralstonia solanacearum ATCC BAA 1114
Lelliottia amnigena ATCC 33072
Klebsiella quasipneumoniae ATCC 700603
Proteus mirabilis ATCC 12453
Shigella flexneri ATCC 25929
Glaesserella parasuis ATCC 19417
Klebsiella pneumoniae subsp pneumoniae ATCC 29665
Corynebacterium glutamicum ATCC 13032
Vibrio diabolicus ATCC 33840
Aggregatibacter actinomycetemcomitans ATCC 29522
Candida krusei ATCC 28870
Serratia marcescens ATCC 14756
Legionella feeleii ATCC 35072
Acinetobacter baumannii ATCC 15308
Actinomyces viscosus ATCC 15987
Priestia megaterium ATCC 9885
Candida albicans ATCC 10231
Acinetobacter calcoaceticus ATCC 23055
Edwardsiella tarda ATCC 15947
Enterococcus faecalis ATCC 700802
Bacillus subtilis ATCC 23857
Bifidobacterium breve ATCC 15700
Shigella flexneri ATCC 12022
Staphylococcus aureus subsp aureus ATCC 27698
Escherichia coli ATCC BAA 2340
Streptococcus salivarius subsp salivarius ATCC 13419
Staphylococcus epidermidis ATCC 35983
Klebsiella oxytoca ATCC 13182
Listeria seeligeri ATCC 35967
Streptococcus dysgalactiae subsp dysgalactiae ATCC 43078
Mycobacterium tuberculosis ATCC 35838
Chromobacterium violaceum ATCC 12472
Serratia marcescens ATCC 14041
Acinetobacter baylyi ATCC 33304
Bacillus paranthracis ATCC 10702
Vibrio alginolyticus ATCC 17749
Enterococcus faecium ATCC BAA 2316
Enterococcus faecalis ATCC 51575
Streptococcus agalactiae ATCC 12386
Mycobacterium smegmatis ATCC 19420
Clostridium septicum ATCC 12464
Lactobacillus reuteri ATCC 23272
Escherichia coli ATCC 23631
Staphylococcus xylosus ATCC 29971
Proteus mirabilis ATCC 7002
Lactobacillus acidophilus ATCC 53544
Enterococcus hirae ATCC 10541
Porphyromonas gingivalis ATCC 33277
Staphylococcus aureus subsp aureus ATCC 49230
Lacticaseibacillus paracasei ATCC 334
Salmonella enterica subsp enterica serovar Pullorum ATCC 9120
Streptococcus gallolyticus ATCC 49147
Staphylococcus saprophyticus ATCC 49453
Escherichia coli ATCC BAA 2523
Enterococcus faecalis ATCC 7080
Enterococcus faecalis ATCC 51299
Staphylococcus aureus ATCC BAA 1556
Salmonella enterica subsp enterica serovar Typhi ATCC 19430
Mammaliicoccus sciuri ATCC 29062
Bifidobacterium longum subsp longum ATCC 15707
Saccharomyces cerevisiae ATCC 7754
Mycobacterium smegmatis ATCC 700084
Issatchenkia orientalis ATCC 6258
Clostridium perfringens ATCC 13124
Clostridium perfringens ATCC 12915
Pseudomonas paraeruginosa ATCC 9027
Staphylococcus saprophyticus subsp saprophyticus ATCC 15305
Pseudomonas aeruginosa ATCC 35151
Klebsiella aerogenes ATCC 13048
Escherichia coli ATCC BAA 3054
Penicillium chrysogenum ATCC 10106
Vibrio harveyi ATCC 14126
Cronobacter sakazakii ATCC 29544
Staphylococcus aureus ATCC 33592
Listeria monocytogenes ATCC 19111
Clostridioides difficile ATCC BAA 1382
Bacillus subtilis ATCC 6051
Kluyveromyces marxianus ATCC 2512
Acinetobacter baumannii ATCC BAA 747
Mycobacterium tuberculosis ATCC 35822
Proteus vulgaris ATCC 29905
Klebsiella aerogenes ATCC 35029
Klebsiella pneumoniae subsp pneumoniae ATCC 15380
Bacillus cereus ATCC 49064
Mycobacterium chelonae ATCC 35752
Micrococcus yunnanensis ATCC 7468
Escherichia coli ATCC 43895
Trichophyton mentagrophytes ATCC 28146
Cronobacter muytjensii ATCC 51329
Aspergillus ustus ATCC 1041
Vibrio vulnificus ATCC 29307
Aspergillus niger ATCC MYA 4892
Porphyromonas gingivalis ATCC 49417
Listeria monocytogenes ATCC BAA 751
Vibrio cholerae ATCC 14035
Staphylococcus aureus ATCC 10832
Escherichia coli ATCC BAA 3051
Vibrio parahaemolyticus ATCC 33847    
\end{lstlisting}

\noindent
\textbf{All the genome from NCBI:}
\begin{lstlisting}
Escherichia coli MG1665: GCF_000005845.2
Escherichia coli UMNK88: GCF_000212715.2
Pseudomonas aeruginosa PA14: GCF_045689255.1
Staphylococcus aureus USA300: GCF_000013465.1
\end{lstlisting}

\section{Strain knowledge searching prompt}\label{secA1}

\begin{lstlisting}[language=Python]
model="qwen-max-0125",  
messages=[
{'role': 'system', 'content': 'You are a scientist working on Antibiotic resistance.'},
{'role': 'user', 'content': f"""Your task is to provide a concise and informative description of the bacterial strain {strain_name}, including:
    
    1. Species Information: Identify the species to which this strain belongs, specifying whether it is Gram-positive, Gram-negative, Fungi, Archaea, or Protozoa. Describe its notable physiological traits.
    2. Unique Mutations: Describe any distinctive genetic mutations identified in this strain compared to the wild-type strain of the same species, particularly those affecting virulence factors, metabolic pathways, or the plasma membrane. Explain how these mutations modify the bacterium's behavior, physiology, or pathogenicity, and how they may contribute to the development of antimicrobial resistance.
    3. Antibiotics and antimicrobial peptides Resistance: Outline any known antibiotics and antimicrobial peptides resistance mechanisms associated with this strain, including specific resistance genes or mutations. Describe the molecular mechanisms by which these confer resistance to particular antibiotics. Make sure to include the corresponding MIC value if you can find it. If you can't find corresponding MIC and molecular mechanisms, don't mention anything about that and keep your response as concise as you can.
    4. Antibiotics and antimicrobial peptides Sensitivity: Identify antibiotics and antimicrobial peptides to which this strain is known to be sensitive. Explain the mechanisms by which these antibiotics exert their effects on the strain. Make sure to include the corresponding MIC value if you can find it. If you can't find corresponding MIC and molecular mechanisms, don't mention anything about that and keep your response as concise as you can.
    In your response, do not provide a summary; instead, list the information as separate points as follows:
    Species Information: ...
    Unique Mutations: ...
    Antibiotic and antimicrobial peptides Resistance: ...
    Antibiotic and antimicrobial peptides Sensitivity: ...
    References: ... (This is just a place holder section, you MUST Ensure that the description is based on current scientific knowledge and has a reference but don't include this section in your response)
    
    Important Instructions:
    1. Ensure that the description is based on current scientific knowledge and includes relevant references where applicable, do not insert references before the Reference section.
    2. Although the strain ID I provided is from ATCC, the same strain may also be cataloged under different identifiers in other databases, such as DSM, KCTC, NCTC, JCM, or other unique numbering systems beyond these examples. Please make sure to cross-reference these alternative ID systems when searching for relevant information.
    3. Do not include any bulletin points in your response.
    4. If information about the strain is unavailable or cannot be found, JUST respond with 'None' in the corresponding section! Do not respond with any further explanation!"""}],
        extra_body={
            "enable_search": True
        }
\end{lstlisting}

\end{appendices}

\end{document}